\documentclass[pmlr]{jmlr}% new name PMLR (Proceedings of Machine Learning)

\RequirePackage{graphicx}
 \usepackage{booktabs}
\usepackage{longtable}% for long tables
\usepackage{float}   

\makeatletter
\def\set@curr@file#1{\def\@curr@file{#1}} %temp workaround for 2019 latex release
\makeatother
\usepackage[load-configurations=version-1]{siunitx} % newer version

\theorembodyfont{\upshape}
\theoremheaderfont{\scshape}
\theorempostheader{:}
\theoremsep{\newline}

\jmlrproceedings{PMLR}{Proceedings of Machine Learning Research}
\jmlrvolume{340}
\jmlryear{2026}
\jmlrworkshop{Machine Learning for Healthcare}

\usepackage{pdflscape}
\usepackage[table]{xcolor}         % colors
\usepackage{multirow}
\usepackage{natbib}
\usepackage{booktabs}

\title[Inference-Time Control for Sepsis Treatment with Generative Patient Digital Twins]{EHR-MPC: Inference-Time Control for Sepsis Treatment with Generative Patient Digital Twins}

\author{\Name{Joshua Pickard}% \footnotemark
       \Email{jpickard@broadinstitute.org}\\
       \addr Broad Institute of MIT and Harvard
       \AND
       \Name{Wei Qi}\Email{wqi2@bwh.harvard.edu}\\
       \addr Broad Institute of MIT and Harvard
       \AND
       \Name{Na Li}
       \Email{nali@seas.harvard.edu}\\
       \addr Harvard University
       \AND
       \Name{Ann Woolley}
       \Email{awoolley@bwh.harvard.edu}\\
       \addr Brigham and Women’s Hospital
       \AND
       \Name{Lisa Cosimi}
       \Email{lcosimi@bwh.harvard.edu}\\
       \addr Brigham and Women’s Hospital
       \AND
       \Name{Roy Kishony}
       \Email{rkishony@technion.ac.il}\\
       \addr Technion–Israel Institute of Technology
       \AND
       \Name{Deborah Hung}
       \Email{hung@molbio.mgh.harvard.edu}\\
       \addr Broad Institute of MIT and Harvard
} 

\usepackage{algorithm}
\usepackage{algorithmic}

\begin{document}

\maketitle

% \footnotetext[0]{THANK YOU}

\vspace{-10mm}
% \begin{center}
%     \textcolor{red}{\textbf{Please adjust/reorder authors as you see fit or add emails.}}
% \end{center}

\begin{abstract}
Sepsis is a leading cause of mortality, yet optimal treatment policies remain contested.
Existing reinforcement learning (RL) approaches learn fixed strategies for sepsis treatment, limiting adaptability to changing clinical objectives during inference.
We propose \textsc{EHR-MPC}, a framework that decouples learning patient dynamics from optimizing treatment by training a patient digital twin in the form of a generative electronic health record (EHR) model.
The digital twin predicts clinical trajectories under interventions and enables model predictive control (MPC) to optimize treatments via inference-time planning over simulations.
We evaluate \textsc{EHR-MPC} on a multicenter ICU sepsis cohort spanning 8 hospitals in the Mass General Brigham health system using both off-policy importance sampling and on-policy simulation-based evaluation.
Relative to RL baselines, \textsc{EHR-MPC} achieves comparable off-policy performance and improved simulation performance.
Unlike RL, this work frames sepsis treatment optimization as inference-time control over learned patient dynamics, establishing a general framework for decision making with generative clinical models.
\end{abstract}

% \begingroup
% \renewcommand{\thefootnote}{\fnsymbol{footnote}}
% \footnotetext[1]{The authors thank the Eric and Wendy Schmidt Center (EWSC) and Center for Integrated Solutions to Infectious Diseases (CISID) at the Broad Institute of Harvard and MIT for support.}
% \endgroup

\section{Introduction}
Treatment of sepsis patients is a major healthcare challenge affecting more than 48 million people and resulting in 11 million deaths per year globally \citep{who2024sepsis}.
Extensive work has explored optimal sepsis treatment policies in both clinical trials \citep{venkatesh2018adjunctive, annane2018hydrocortisone} and reinforcement learning \citep{raghu2017deep, raghu2017continuous, raghu2018model, komorowski2018artificial, huang2022reinforcement}.
In particular, corticosteroid administration remains a longstanding clinical question, with more than 60 randomized controlled trials conducted \citep{schumer1976steroids, annane2025corticosteroids}.
Still, evidence guiding which patients should receive corticosteroids, as well as the appropriate timing and dosage, remains contested \citep{marik2018steroids}.
% Still, evidence guiding which patients should receive corticosteroids --- and when and at what dosage --- remains contested \citep{marik2018steroids}.

While both clinical trials and reinforcement learning (RL) aim to identify optimal policies for sepsis management, their recommendations remain largely disconnected.
For instance, one RL study suggests that “optimal treatment [with corticosteroids] may be more restrictive than routine clinical practice” \citep{bologheanu2023development}, in contrast to randomized clinical trial evidence indicating that corticosteroids “probably reduce 28-day and hospital mortality among patients with sepsis” \citep{annane2025corticosteroids}.
This discrepancy highlights a limitation of RL approaches, as despite strong retrospective performance under off-policy evaluation, learned policies are often difficult to interpret, adapt, or validate in clinical practice \citep{zhang2022interpretable, frommeyer2025reinforcement}.
Optimizing a fixed reward function without explicitly modeling patient dynamics limits the ability of RL-based treatment strategies to accommodate individualized clinical goals or to adapt to competing objectives \citep{jayaraman2024primer}.
In practice, sepsis management requires balancing trade-offs such as short-term hemodynamic stabilization versus long-term organ-system preservation, informed by clinician judgment, evolving standards of care, and patient-specific context \citep{prescott2026surviving}.
Because RL typically entangles patient dynamics and reward specification within a single model, adapting a learned policy to new clinical objectives requires retraining, limiting robustness to changing goals, deployment settings, and human-in-the-loop constraints.

% However, because conventional RL frameworks entangle implicit patient dynamics and clinical rewards within a single model, adapting RL policies to new clinical objectives typically requires retraining, limiting their suitability for human-in-the-loop less robust to changes in objectives and deployment context.

\begin{figure}[t]
    \centering
    \includegraphics[width=\linewidth]{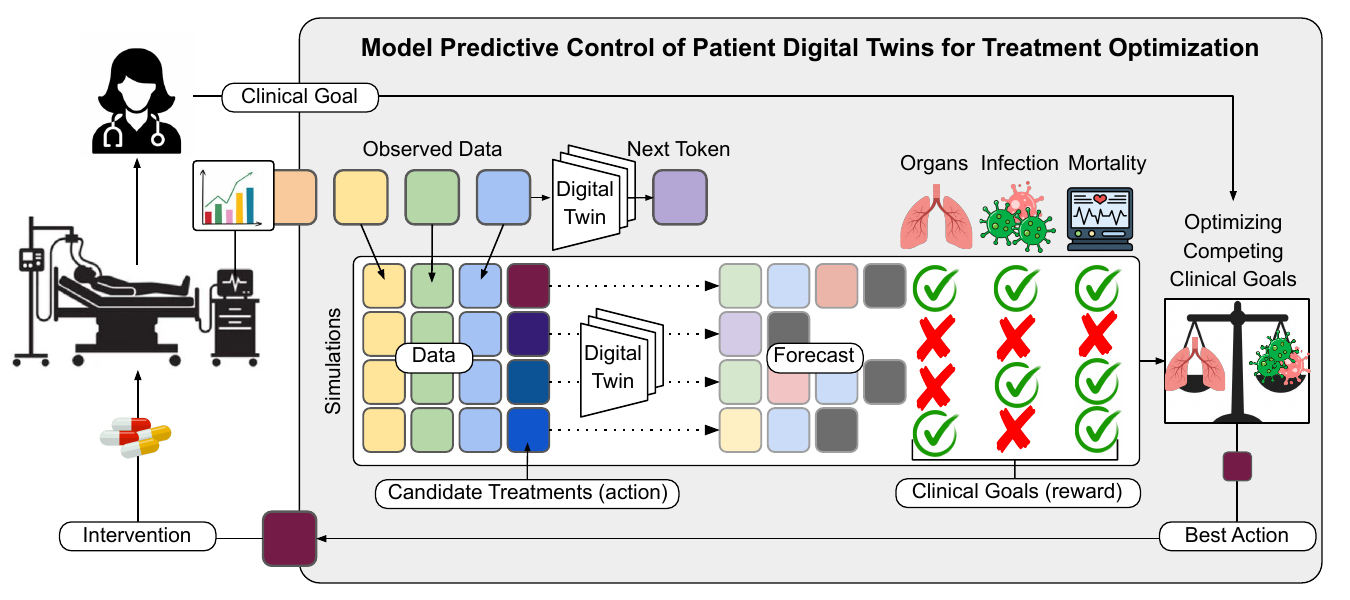}
    \vspace{-10mm}
    \caption{\small\textbf{Digital Twin MPC for Treatment Optimization.} (left) A patient generates data that is observed by a clinician, tokenized, and sent to the digital twin. (simulation) Using observed data, a digital twin performs forecasts on the state of the patient as if several candidate treatments were applied. Each forecast is evaluated according to several clinical outcomes. (optimization) Based on clinical goals defined at inference time, the action token generating the best simulated outcome is translated to the proposed intervention.}
    \label{fig:main}
\end{figure}

The need to improve sepsis management, together with the limitations of prior RL approaches, motivates decoupling the learning of patient dynamics from the optimization of treatment decisions, so that clinical objectives are specified at inference time rather than embedded in a fixed policy.
We propose \textsc{EHR-MPC}, a framework that learns a generative digital twin of patient trajectories from electronic health record (EHR) data and performs treatment optimization using model predictive control (MPC).
The digital twin simulates counterfactual trajectories under candidate interventions, while an MPC controller evaluates and selects action sequences according to clinically specified objectives at inference time, forming a closed-loop system that can incorporate clinician feedback (Fig.~\ref{fig:main}).
This framework enables treatment strategies to be optimized under new objectives without retraining, supporting flexible, objective-aware, and human-in-the-loop clinical decision-making.

% The outstanding need for improving sepsis management, together with the limitations of prior RL approaches, motivates a separation between patient modeling and decision-making, where clinical objectives are no longer embedded within a single fixed policy but are instead evaluated over a learned model of patient dynamics.
% %
% We propose \textsc{EHR-MPC}, a framework that learns a generative digital twin of patient trajectories from electronic health record (EHR) data and performs treatment optimization at inference time using model predictive control (MPC).
% %
% As illustrated in Fig.~\ref{fig:main}, the digital twin simulates counterfactual patient trajectories under candidate interventions, which are then evaluated by a controller that searches over action sequences according to clinically specified objectives, forming a closed-loop system that can incorporate feedback from clinicians.
% %
% This framework separates learning patient dynamics from optimizing treatment decisions, enabling interventions to be re-evaluated under different objectives without retraining the underlying model.
% %
% By shifting policy optimization from training to inference time, \textsc{EHR-MPC} provides a flexible and extensible framework for treatment selection that supports interactive and objective-aware clinical decision-making.

\vspace{2mm}
\noindent
\textbf{Generalizable Insights about Machine Learning in Healthcare.}
This work suggests three potential advantages of decoupling dynamics from policy optimization.
First, separating patient dynamics from intervention policies changes the structure of the learning problem, as modeling patient trajectories explicitly yields a digital twin that is reusable across treatment objectives, whereas policy learning must entangle both dynamics and treatment objectives.
Second, generative EHR models support decision-making algorithms, where pretrained patient digital twins enable downstream optimization via inference-time methods such as model predictive control (MPC).
This suggests a broader class of clinical machine learning systems that extend beyond prediction to planning over learned patient simulators.
Third, moving from training-time policy optimization to inference-time planning enables adaptation to changing clinical objectives without retraining, supporting more flexible and objective-aware decision-making.
Together, these insights point toward a paradigm in which reusable models of patient dynamics replace fixed policies as the primary object of learning, with treatment decisions obtained through inference-time control over these models.

% \textbf{Generalizable Insights about Machine Learning in the Context of Healthcare.}
% This work highlights three generalizable insights.
% %
% First, decoupling patient dynamics from intervention policies changes the learning objective of treatment optimization. Modeling patient trajectories provides a more stable and reusable representation than directly learning treatment policies, which must implicitly encode both dynamics and objectives.
% %
% Second, generative EHR models can serve as substrates for decision-making, where pretrained patient digital twins support downstream treatment optimization via methods including MPC. This defines a new class of tasks that extend beyond prediction to inference-time planning over clinical foundation models.
% %
% Third, shifting optimization from training-time policy learning to inference-time planning enables flexible and objective-aware decision-making, allowing treatment strategies to be adapted to new clinical goals without retraining the underlying model.
% %
% Collectively, these insights suggest a move away from fixed policy learning toward reusable models of patient dynamics that can be queried and controlled to support adaptive clinical decision-making.

\section{Related Work}

We compare two existing paradigms for resolving the question of steroid use in sepsis and introduce a third framework. Table~\ref{tab: comparison} summarizes this section.

\paragraph{Optimal use of steroids in sepsis remains unresolved.}
The use of corticosteroids for sepsis and septic shock has been debated for decades, with trials dating back to the 1970s \citep{schumer1976steroids}.
Despite substantial randomized controlled trials, no consensus has emerged on if, when, and for whom, steroids are beneficial \citep{venkatesh2018adjunctive, annane2018hydrocortisone, annane2025corticosteroids}.
This ambiguity reflects that steroid therapy involves a trade-off between suppressing harmful inflammation and impairing host immune response, and the balance of these effects varies across patients and over time \citep{wiersinga2018handbook}.
As a result, the optimal treatment strategy is dynamic and patient-specific, depending on the patient state and treatment goals.
These challenges make steroid use in sepsis management a longstanding, open challenge and an example of sequential decision-making under uncertainty, motivating the use of machine learning to find improved treatment strategies.

\paragraph{Reinforcement learning for sepsis.}
The AI Clinician was an RL model trained to recommend vasopressor and fluid dosing for sepsis patients, reporting improved off-policy evaluation performance \citep{komorowski2018artificial}.
Subsequent work expanded the action space to include steroids \citep{bologheanu2023development}, antibiotics \citep{futoma2018learning, wang2024clinical}, and continuous treatment representations \citep{raghu2017continuous, li2020optimizing}, while also incorporating safety constraints and uncertainty modeling \citep{nanayakkara2022unifying, tu2025offline}.
A predominant RL paradigm has relied on Q-learning–based methods, with policy evaluation commonly performed using off-policy estimators such as importance sampling or doubly robust methods \citep{peng2018improving, jia2020safe, liu2021offline, huang2022reinforcement, bock2022superhuman, wu2023value, zhang2024optimizing, choi2024deep, drudi2024reinforcement}.
Across these approaches, policies are learned from retrospective EHR data by optimizing value functions that implicitly encode both patient dynamics and clinical objectives.
Despite extensive methodological development, these approaches have had limited translation into clinical practice.
More broadly, many results are validated on a small number of benchmark datasets, such as MIMIC and eICU, raising concerns about generalizability \citep{pollard2018eicu, johnson2023mimic}.
Collectively, this suggests that learning policies directly from observational data without explicitly modeling patient dynamics can yield brittle and potentially misaligned treatment recommendations.

\begin{table}[t]
\centering
\small
\setlength{\tabcolsep}{8pt}
\caption{\small \textbf{Comparison of paradigms for sepsis treatment optimization.} The EHR-MPC framework combines the interpretability  of clinical trials with the data-driven approach of RL and enables  inference-time adaptation. Characterizations vary by implementation.}\label{tab: comparison}
\vspace{-3mm}
\begin{tabular}{l|cc>{\columncolor[gray]{0.92}}c}
\toprule
 & \textbf{Clinical Trials} & \textbf{RL (value-based)} & \textbf{EHR-MPC (ours)} \\
\midrule
Data           & Prospective      & Retrospective    & Retrospective     \\
Interpretability & High           & Limited          & High              \\
Policy         & Fixed protocol   & Trained       & Inference-time    \\
Dynamics       & Not modeled      & Implicit         & Explicit          \\
Adaptivity     & Limited          & Limited          & High              \\
Evaluation    & Randomization    & Off-policy & Simulation        \\
\bottomrule
\end{tabular}
\end{table}

\paragraph{Foundation models as patient digital twins for control.}
Large language models trained on clinical notes and structured EHR tokens have demonstrated strong performance on diagnostic and predictive tasks \citep{lee2020biobert, rasmy2021med}.
More recently, trajectory-level foundation models have been shown to capture temporal physiological structure, enabling prediction of patient evolution over time~\citep{renc2024zero, renc2025foundation, makarov2025large, li2025mira}.
These models can be interpreted as \emph{digital twins}, that is, data-driven simulators of patient dynamics that evolve under different clinical interventions.
This framing enables decision-making through optimization over predicted patient trajectories \citep{alge2024continuous, pickard2025dynamic, prunella2026evolutionary}.
In parallel, classical model predictive control (MPC) has been applied in clinical settings such as drug delivery and glucose regulation, but relies on hand-crafted physiological models that limit flexibility and scalability \citep{ionescu2008robust, bruttomesso2009closed, nacscu2014advanced}.
In contrast, learned digital twins provide a data-driven foundation for MPC, enabling inference-time optimization over 
treatments to improve patient outcomes.

\section{EHR-MPC: Electronic Health Record Model Predictive Control}
\label{sec:methods}

We introduce \textsc{EHR-MPC}, a framework that decouples (i) learning patient dynamics from real-time EHR data and (ii) optimizing clinical decisions.
We formalize drug administration as a sequential decision-making problem, then present a two-stage approach of a digital twin patient model and an MPC controller that optimizes treatment actions during inference.

\subsection{Problem Formulation}
We model sepsis management as a state--action--reward process.
At each decision time $t$, the \emph{state} $x(t)$ consists of all clinical data available in the ICU (vitals, labs, interventions, demographics, and device settings).
An \emph{action} $a(t)$ includes any clinical intervention, such as the administration of corticosteroids.
A \emph{reward} function $R(x(t), a(t))$ encodes the clinical objective, such as the Sequential Organ Failure Assessment (SOFA) score or mortality.

This formulation is compatible with both previous RL studies and the proposed framework.
Unlike RL, which explicitly trains a policy by entangling patient dynamics and clinical objectives during training, EHR-MPC trains only a generative model of patient dynamics and obtains treatment decisions entirely through inference-time planning.

\subsection{Learning and Simulating Patient Digital Twins}
We learn a model to forecast patient trajectories by training on tokenized EHR data.
Data are pulled directly from an operational EHR database shared by several hospitals so that the model is designed for constraints that arise during real-time deployment in a clinical setting.

% designed for constraints that arise during the real-time deployment at a clinical setting.

% \paragraph{Tokenized EHR.}
% Each patient encounter is represented as a sequence of tokens:
% \begin{equation}
%     x(t_k)_{1:k} = (x_1, x_2, \ldots, x_k),
% \end{equation}
% where tokens are ordered in time and correspond to clinical events such as measurements of vitals, labs, device settings, or drug administrations.
% The state variable $x_{1:k}$ represents both historic patient data and clinical interventions.
%
% Continuous measurements are discretized into bins, and interventions (e.g., administering corticosteroids) are represented with action tokens.
%
% Additional special tokens to indicate start and end of the sequence, unknown clinical events, and the passing of four hour intervals, are also included.

% With tokens $x_{1:t_k},$ the patient state is updated to time $t+h$ via a streaming update operator
% \begin{equation}
% x_{1:(t+h)_k} \leftarrow \mathrm{StreamUpdate}(x_{1:t_k},\ t+h),
% \end{equation}
% where $h$ is a time horizon and $(t+h)_k$ is the token index for all events occuring up until time $t+h.$
% Tokens are indexed by event order $k$ and ordered by time $t$ to accomidate the asynchronous and irregular observation schedules. This is compatible with event driven control notation
% The StreamUpdate operator, which generates new tokens from real-time EHR system is realized by learning a generative patient digital twin.

% \begin{equation}
%     x_{1:k} = (x_1, x_2, \ldots, x_k),
% \end{equation}

\paragraph{Tokenized EHR.}
Each patient is represented as a sequence $x_{1:k} = (x_1, x_2, \ldots, x_k)$ where tokens $x_i$ are ordered in time and correspond to clinical events such as measurements of vitals, labs, device settings, or drug administrations. At time $t$ the state is $x(t)=x_{1:k_t}.$
The state variable $x_{1:k}$ represents both historic patient data and clinical interventions.
Continuous measurements are discretized into bins, and interventions (e.g., administering corticosteroids) are represented with action tokens.
Additional special tokens to indicate start and end of the sequence, unknown clinical events, and the passing of four hour intervals, are also included.
As an example, a patient's EHR data over a 4-hour interval may be encoded as the token sequence
\begin{center}
\small
\texttt{
[TIME\_4H], [HR\_60-65], [LACTATE\_2.1-2.3], [HYDROCORTISONE\_IV], ..., [TIME\_4H],
}
\end{center}
where the tokens encode the observation time, discretized heart rate and serum lactate measurements, intravenous hydrocortisone administration, and other clinical events.
From time $t$ to time $t+h,$ the patient state updates as:
\begin{equation}
    x_{1:k_{t+h}}\leftarrow\mathrm{StreamUpdate}(x_{1:k_t}, t+h).
\end{equation}
There are $k_{t+h}-k_t$ new tokens added to the patient state during this time interval.
The duration $h$ is a time horizon that represents intervals such as the time delay in streaming data from the EHR database or the forecasting horizon for predicting patient outcomes.
Tokens are indexed by both order ($k$) and time ($t$) to accommodate irregular sampling schedules in EHR and event triggered control \citep{heemels2012introduction}.
The {StreamUpdate} operator is modeled by the generative patient digital twin.

\paragraph{Simulated patient dynamics.}
Given the current patient history $x_{1:k_t}$, the digital twin defines the next-token distribution
\begin{equation}
    p_\theta(x_{k_t+1} \mid x_{1:k_t}),
\end{equation}
where $\theta$ are the parameters of the twin model.
This distribution $p_\theta$ can be instantiated using sequence models such as transformers \citep{makarov2025large, li2025mira}.
We train the patient digital twin using a GPT-2-style causal language model architecture, similar to ETHOS \citep{renc2024zero}, although the approach readily extends to larger pretrained models and alternative architectures.
% We train the patient digital twin with a GPT-2-style causal language model architecture, similar to ETHOS \citep{renc2024zero}, though the formulation readily scales to larger pretrained models and alternative model architectures.
% We train a moderate-scale model on our cohort, though the formulation readily scales to larger pretrained models.

The digital twin $p_\theta$ is trained to forecast patient trajectories over a fixed horizon $h$ (e.g., the next 24 hours).
Beginning from time $t$, this corresponds to generating tokens up to a horizon $t+h,$ which are distributed according to
\begin{equation}
    p_\theta(x_{k_t:k_{t+h}}\ |\ x_{k_t})=\prod_{i=k_t}^{k_{t+h}}p_\theta(x_i\ |\ x_{<i}).
\end{equation}
% In practice, the goal is to forecast patient trajectories over a fixed horizon (e.g., the next 24 hours).
%
% Starting from time $t$, this corresponds to generating tokens up to a horizon $t+h$ via
% \begin{equation}
%     p(x_{k:k+\Delta(t,h)}) = \prod_{i=k}^{k+\Delta(t,h)}p(x_{i}|x_{<i}).
% \end{equation}
% The number of tokens $\Delta(t,h)$ required to reach horizon $h$ is not known a priori and depends on the realized sequence.
%
The number of tokens required to reach time $t+h$ depends on the realized sequence.
For example, when $h=24$ hours, generation proceeds until six $\texttt{[4 hour]}$ tokens have been produced.

\paragraph{Simulation via Next-Token Prediction.}
To generate patient trajectories under a candidate treatment sequence $a_{k_t:k_{t+h}}$, we \emph{force} the corresponding action tokens into the input and recursively generate future tokens using the autoregressive model:
\begin{equation}\label{eq: action forced distribution}
    p(x_{k_t:k_{t+h}} \mid x_{1:k_t}, a_{k_t:k_{t+h}}) = \prod_{i=k_t}^{k_{t+h}} p(x_i \ |\ x_{<i}, a_{<i}),
\end{equation}
% \begin{equation}\label{eq: action forced distribution}
%     p(x_{k+1:k+\Delta(t,h)} \mid x_{1:k}, a_{t:t+h}) = \prod_{i=k}^{k+\Delta(t,h)} p(x_i \mid x_{<i}, a_{t:t_i}),
% \end{equation}
where $a_{k_t}$ denotes the actions applied up to step $i$.
The resulting model defines a controlled generative process \citep{plumerault2020controlling}.
This procedure yields a counterfactual trajectory in which the generated tokens represent the model’s predicted evolution of patient state (e.g., vitals, labs, and interventions) under the proposed treatment.

Eq.~\ref{eq: action forced distribution} enables the model $p_\theta$ to act as a \emph{patient digital twin}: a data-driven, virtual replica that evolves in response to real-time EHR data and proposed interventions \citep{laubenbacher2024digital}.
The digital twin can be viewed as a simulator of patient trajectories, forecasting how physiology may evolve under alternative treatment strategies and enabling comparison of candidate interventions before action.

\subsection{Model Predictive Control via Token-Forced Rollouts}
To derive treatment recommendations, we apply model predictive control (MPC) to the learned patient digital twin $p_\theta$ at each decision time. Rather than learning a fixed policy, the controller performs explicit planning at inference time.
MPC evaluates candidate intervention sequences by simulating their effect on future patient trajectories, scores the trajectories under a given objective, and selects the action sequence with the highest predicted utility.
Because MPC operates entirely at inference time, the clinical objective (i.e., reward function) can be specified after training. This enables flexible optimization across different and potentially evolving clinical goals, and allows the controller to evaluate interventions that were not explicitly anticipated during training of the digital twin.

\paragraph{Token-forced rollouts.}
Given a candidate action sequence $a_{t:t+h}$ over a horizon $h$, we simulate a counterfactual trajectory by \emph{forcing} the corresponding action tokens into the model input and generating future tokens by sampling from the distribution defined in Eq.~\ref{eq: action forced distribution}.
At each step $k_t$, the next token is sampled
\begin{equation}\label{eq:sim}
    \hat{x}_{k_{t+1}} \sim p(x_{k_{t+1}}\ |\  x_{1:k_t}, a_{k_t:k_{t+h}}),
\end{equation}
yielding a trajectory $\hat{x}_{t:t+h}$ that represents the predicted evolution of patient state under the proposed interventions.

This procedure simulates ``what would happen'' under a treatment sequence.
By explicitly modeling intervention effects, the framework yields (i) a mechanistic forecast of patient trajectory evolution, rather than only final outcomes, and (ii) a basis for evaluating and comparing outcomes across alternative treatment strategies.

\paragraph{Trajectory evaluation.}
Each simulated trajectory is evaluated according to a reward function.
The purpose of these functions is to evaluate how well each patient trajectory fits defined clinical endpoints.
Common endpoints considered in clinical trials for learning improved sepsis treatment policies include: patient mortality, length of stay in the hospital and ICU, and organ failure, which is commonly measured with the Sequential Organ Failure Assessment (SOFA) score \citep{vincent1996sofa}.

A reward function $R:x_{1:k_t}\xrightarrow{}\mathbb{R}$ assigns a numeric score to each patient trajectory.
Computationally, $R$ acts as a decoding objective over trajectories, converting predicted clinical sequences into a scalar utility that can be optimized via search over intervention candidates.
These reward functions can be specified in two ways.
In an explicit formulation, specific tokens (e.g., \texttt{[Mortality]}) directly induce reward or penalty.
In an implicit formulation, clinically relevant outcomes such as SOFA score or length of stay are estimated from the trajectory rather than directly observed in the token sequence.
This distinction is necessary because several endpoints are not directly represented as tokens.
For example, length of stay is only known after transfer or discharge, and SOFA components may be only partially and irregularly observed during an ICU stay.
As a result, these quantities are partially observable and must be inferred from the evolving trajectory rather than read off the sequence directly.
In both formulations, constructing $R$ can be achieved for any clinical objective by constructing either explicit functions operating on the token sequence or with task-specific heads that predict clinical outcomes.

Evaluation is performed at inference time, enabling the same learned dynamics model to support optimization over multiple, potentially changing objectives without retraining $p_\theta$.
This flexibility is a consequence of decoupling patient dynamics from clinical objectives.
Because $R$ is never embedded in the model $p_\theta$, new reward functions can be introduced or modified at deployment time without any retraining.
This allows a single trained digital twin to simultaneously serve clinicians with different treatment priorities, adapting to evolving standards of care or patient-specific goals without additional model development.

\paragraph{Action selection and receding-horizon control.}
During inference, the controller selects a sequence of interventions by solving a planning problem over the digital twin:
\begin{equation}\label{eq: optimization}
a_t^* = \arg\max_{a_{k_t:k_{t+h}}}
\ \mathbb{E}_{\hat{x}_{k_{t}:k_{t+h}} \sim p_\theta(\cdot \mid x_{1:k_t}, a_{k_t:k_{t+h}})}
\left[ R(\hat{x}_{k_t:k_{t+h}}) \right].
\end{equation}
In practice, this optimization is intractable to solve exactly and is approximated
via simulation-based search \citep{garcia1989model}.
Candidate action sequences are sampled (or constructed

\vspace{-0.2mm}
\noindent
\begin{minipage}[t]{0.52\textwidth}
via heuristic exploration), their corresponding trajectories are generated by
sampling from $p_\theta$, and the resulting trajectories are scored using $R$.
The action sequence with highest estimated value is selected.
Only the first action $a_t^*$ is executed, after which the system observes updated
patient data and repeats the optimization. This receding-horizon procedure enables
continual re-planning as new information becomes available, analogous to clinical practice
\end{minipage}%
\hfill
\begin{minipage}[t]{0.44\textwidth}
\vspace{-5mm}
\begin{algorithm}[H]
\caption{EHR-MPC}
\label{alg:mpc}
\begin{algorithmic}[1]
\small
\FOR{each decision time $t$}
\STATE Generate candidate treatment sequences
\STATE Simulate patients with $p_\theta$ (Eq.~\ref{eq:sim})
\STATE Select optimal sequence (Eq.~\ref{eq: optimization})
\STATE Apply first treatment action
\STATE Observe updated patient state
\ENDFOR
\end{algorithmic}
\end{algorithm}
\end{minipage}

\vspace{0.7mm}
\noindent 
 where treatment decisions are repeatedly updated in response to evolving
patient state. Algorithm~\ref{alg:mpc} provides a high-level overview of the MPC procedure, while Section~\ref{si:mpc-details} provides the EHR-MPC implementation details, including trajectory simulation and optimization steps.

\paragraph{Methodological interpretation.}
\textsc{EHR-MPC} decouples learning of patient dynamics from decision-making by using a generative digital twin to explicitly simulate future trajectories under candidate intervention sequences.
At inference time, decisions are obtained through simulation-based search, where each candidate intervention is evaluated by rolling out the learned model and scoring the resulting trajectory against a clinically specified objective.
This shifts computation from offline policy fitting to online planning, enabling clinical objectives to be modified without retraining.
This formulation enables three capabilities absent from standard RL approaches, namely (i) inference-time objective specification, (ii) direct inspection of counterfactual patient trajectories, and (iii) reuse of a single dynamics model across multiple clinical goals.

% \paragraph{Methodological interpretation.}
% \textsc{EHR-MPC} decouples learning of patient dynamics from decision-making by using a generative digital twin to explicitly simulate future trajectories under candidate intervention sequences.
% %
% At inference time, decisions are obtained through simulation-based search over actions, where each candidate intervention is evaluated by simulating trajectories with the learned model and scoring the resulting trajectories according to a clinical goal.
% %
% This shifts computation from offline policy fitting to online planning, enabling objectives to be modified without retraining.
% %
% This inference time computation increases flexibility in the online optimization by introducing the capability to directly interrogate generated counterfactual trajectories.
% flexibility in optimization 
% The trade-off is increased inference-time computation due to repeated rollouts, but this is offset by flexibility in optimization and the ability to directly interrogate predicted counterfactual trajectories.

\section{Cohort Selection}

\subsection{Clinical Setting and Data Source}
We assembled a multi-site cohort of ICU patients from a large health system comprising eight hospitals.
This included patient encounters at two academic medical centers (AMC) and six community hospitals (CH) observed from 2022 onward.
Data originate from the institutional electronic health record system, which feeds into a relational database.
Patient encounter information, including lab results, vital signs, locations, medication administrations, diagnoses, and other signals were extracted from the production EHR system.

% \footnotetext{To preserve anonymity, specific details of the health system and data access are currently omitted. Additional details are available upon request or in a finalized version.}

\subsection{Cohort Identification}
We constructed the study cohort from all patient encounters with at least one ICU admission across a curated set of more than 30 critical care units, spanning medical, surgical, cardiac, and mixed ICUs across all participating sites.
From this eligible population, we sampled {36,930 patients}, including both sepsis and non-sepsis ICU admissions.
The resulting cohort is temporally uniform over the collection dates and stratified across ICU locations in proportion to their underlying patient volumes.
Including a broad ICU population was intentional, since sepsis may develop during an ICU stay and diagnostic labeling based on ICD codes is known to be imperfect and sensitive to evolving clinical definitions \citep{liu2022accuracy}.
This design reflects deployment conditions in which the model operates over general ICU admissions rather than a pre-filtered diagnostic cohort, and avoids coupling cohort construction to the same coding schemes used to define the target outcome.
Cohort characteristics stratified by site are reported in Table~\ref{tab:cohort}.

% \subsection{Cohort Identification}
% We constructed the study cohort from all patient encounters with at least one ICU admission across a curated set of 30+ critical care units, spanning medical, surgical, neurological, cardiac, and mixed ICUs across all participating sites. Encounters were required to meet a minimum length-of-stay threshold and to visit designated ICU locations.
% %
% From this eligible population, we sampled \textbf{36,930 patients}, including both patients with and without a sepsis diagnosis code. The resulting cohort is temporally uniform and stratified across ICU locations in proportion to their underlying patient volumes.

% Including a broad ICU population was intentional, as sepsis may develop during an ICU stay and diagnostic labeling via ICD codes is known to be imperfect and sensitive to evolving clinical definitions \cite{liu2022accuracy}.
% %
% This design reflects deployment conditions, where the model operates over undifferentiated ICU admissions rather than a pre-filtered diagnostic subset, and avoids coupling cohort construction to the same coding schemes used to define the target clinical outcome.
% % 
% Cohort characteristics stratified by site are reported in Table~\ref{tab:cohort}.

% \begin{landscape}
{
\small
% Table 1 — Cohort characteristics
\begin{table}[t]
\centering
\caption{\small \textbf{Patient cohort.} $36$k ICU patients from eight hospitals, collected between 2022 and 2026.}
\label{tab:cohort}
\vspace{-3mm}
\fontsize{7.8}{10}\selectfont

\begin{tabular}{l>{\columncolor[gray]{0.92}}rrrrrrrrr}
\toprule
%\textbf{Location} & \textbf{Total} & \textbf{A} & \textbf{B} & \textbf{C} & \textbf{D} & \textbf{E} & \textbf{F} & \textbf{G} & \textbf{H} \\
\textbf{Location}&\textbf{Total}&\textbf{MGH}&\textbf{BWH}&\textbf{SLM}&\textbf{NWH}&\textbf{WDH}&\textbf{BWF}&\textbf{CDH}&\textbf{MVH}\\
% Location & Total & A & B & C & D & E & F & G & H \\
\midrule
\quad\textit{Hospital Type}&&AMC&AMC&CH&CH&CH&CH&CH&CH\\
\textbf{Demographics} &  &  &  &  &  &  &  &  &  \\
\quad\textit{Patients} & 36930 (100.0\%) & 13651 & 12291 & 3541 & 2256 & 1849 & 1465 & 1358 & 519 \\
\quad\textit{LOS (hrs)} & 66.2 (27.6–154.6) & 83.0 & 74.3 & 54.7 & 49.0 & 38.5 & 44.9 & 40.8 & 7.8 \\
\quad\textit{Mortality} & 12657 (34.3\%) & 4285 & 4337 & 1370 & 864 & 561 & 494 & 629 & 117 \\
\quad\textit{Female (n)} & 16101 (43.6\%) & 5487 & 5322 & 1654 & 1146 & 841 & 746 & 641 & 254 \\
\quad\textit{Median Age (yrs)} & 67 (55–76) & 66 & 66 & 68 & 72 & 67 & 67 & 68 & 74 \\
\quad\textit{Median BMI} & 26.9 (23.1–31.4) & 27.0 & 26.8 & 27.0 & 25.7 & 27.8 & 26.4 & 27.1 & 26.0 \\
\textbf{SOFA Scores} &  &  &  &  &  &  &  &  &  \\
\quad\textit{SOFA IQR} & 2–5 & 2–5 & 2–5 & 2–5 & 1–4 & 2–5 & 1–5 & 1–4 & 1–3 \\
\quad\textit{Days $>3$ subscores (\%)} & 78.8 & 79.5 & 81.3 & 77.3 & 70.6 & 74.9 & 75.5 & 64.9 & 57.0 \\
\textbf{Treatments} &  &  &  &  &  &  &  &  &  \\
\quad\textit{Vasopressor} & 21097 (57.1\%) & 9594 & 7067 & 1448 & 1105 & 842 & 572 & 404 & 65 \\
\quad\textit{Antibiotic} & 22635 (61.3\%) & 7430 & 8260 & 2358 & 1386 & 1112 & 967 & 907 & 215 \\
\quad\textit{Corticosteroid} & 12616 (34.2\%) & 4633 & 4405 & 1155 & 722 & 551 & 458 & 567 & 125 \\
\textbf{Diagnoses} &  &  &  &  &  &  &  &  &  \\
\quad\textit{Sepsis} & 4129 (11.2\%) & 1053 & 994 & 660 & 348 & 324 & 296 & 410 & 44 \\
\quad\textit{Septic shock} & 1495 (4.0\%) & 333 & 379 & 211 & 134 & 140 & 135 & 145 & 18 \\
\quad\textit{Respiratory failure} & 6741 (18.3\%) & 1809 & 1865 & 1056 & 353 & 433 & 463 & 667 & 95 \\
\quad\textit{Hepatic failure} & 831 (2.3\%) & 385 & 136 & 110 & 40 & 42 & 51 & 59 & 8 \\
\quad\textit{Heart failure} & 8435 (22.8\%) & 3000 & 2451 & 1069 & 534 & 469 & 409 & 384 & 119 \\
\quad\textit{Coagulopathy} & 1226 (3.3\%) & 406 & 468 & 98 & 81 & 51 & 56 & 43 & 23 \\
\quad\textit{Encephalopathy} & 1277 (3.5\%) & 275 & 222 & 238 & 74 & 130 & 110 & 211 & 17 \\
\quad\textit{Pneumonia} & 5026 (13.6\%) & 1468 & 1113 & 916 & 372 & 304 & 300 & 424 & 129 \\
\quad\textit{Shock (other)} & 3581 (9.7\%) & 1075 & 1396 & 236 & 182 & 246 & 261 & 177 & 8 \\
\bottomrule
\end{tabular}
\end{table}

% \end{landscape}
}

\subsection{Data Extraction}
For each encounter, we assembled a complete longitudinal EHR record by joining across flowsheet, laboratory, medication, procedure, administrative, demographic, and diagnostic tables within the EHR relational database.
We included available records from each patient admission as well as those occurring within two days before or after the patient encounter.
Time-unrestricted historical data for comorbidities, prior diagnoses, mortality outcomes, and admission records to other hospitals were also included.
This design ensures that each patient representation incorporates both proximal clinical dynamics and relevant longitudinal context.

Working with a production EHR database introduces several well-known data quality challenges. First, the database schema evolves over time as EHR systems are updated and hospital configurations change. Second, substantial heterogeneity exists across sites in how clinical variables are recorded, including differences in flowsheet structure, laboratory naming conventions, medication ordering systems, and unit-level documentation practices. Third, many clinical events have uncertain or delayed timestamps, particularly for diagnosis codes, laboratory processing times, and documentation-based observations, which can introduce temporal ambiguity in the recorded trajectories.

These challenges are standard in large-scale EHR analysis and are mitigated through careful manual curation of key mappings and alignment between backend data structures and their corresponding clinical semantics. For example, while the SOFA score is derived from more than 100 raw database fields, these map to a small number of underlying physiological measurements. In our framework, the use of a token-based, self-supervised sequence model provides additional robustness to these issues by learning directly from observed event streams without requiring perfect alignment of individual fields or strict synchronization of measurement times.

\subsection{Tokenized EHR Representation}
Following extraction and preprocessing, each patient trajectory was converted to a discrete token sequence suitable for modeling as described in Section~\ref{sec:methods}.
Following \cite{renc2024zero}, continuous measurements were discretized into bins and medications, procedures, and administrative (admit, transfer, discharge, etc.) events were mapped to dedicated vocabulary entries.
The tokenized dataset contained 469 million content tokens, 90.3\% of which indicated observational data rather than interventions or other data.
The median patient trajectory sequence length contained 20,232 tokens of which 1,058 are unique.
Detailed token-type breakdowns and vocabulary statistics are provided in Section \ref{apdx: tokenization}.

\section{Experiments}
Following our decoupled formulation of learning and policy optimization, experiments are organized into two stages. First, validating the digital twin as a predictive and intervention-aware model (Section~\ref{sec: exp 1}). Second, evaluating treatment policies derived from the model using both off-policy and simulation-based methods (Section~\ref{sec: exp 2}) \footnotemark.

\footnotetext{See the appendix for complete experimental details.}

\subsection{Digital Twin Construction and Forecasting Evaluation}\label{sec: exp 1}
We train a transformer to parameterize $p_\theta$ (Section~\ref{sec:methods}) and evaluate it on the following tasks.

\begin{figure}[t]
    \centering
    \includegraphics[width=\linewidth]{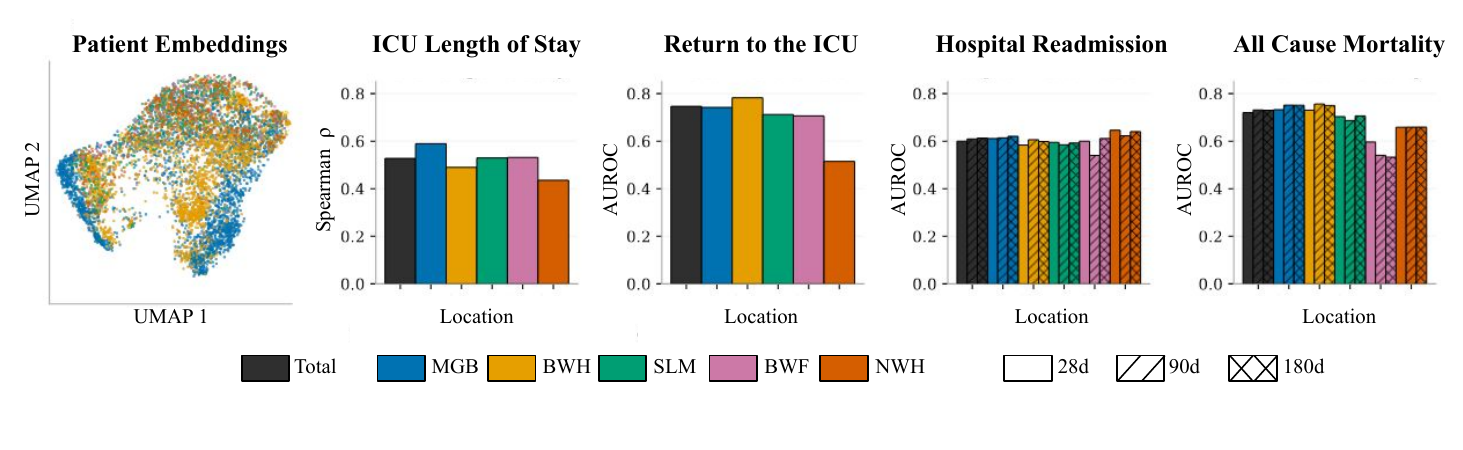}
    \vspace{-15mm}
    \caption{\small\textbf{Representations and Forecasting.} Prediction of outcomes from patient embeddings and random forest classifiers. Model performance is consistent across hospital settings and prediction horizons.}
    \label{fig:R1}
\end{figure}

\paragraph{Outcome Prediction from Learned Patient Representations.}
We assess representation quality learned by the digital twin according to their ability to predict several standard clinical endpoints.
We fit lightweight random forest models to predict the ICU length of stay (time-to-event), ICU readmission within the same encounter (binary), hospital readmission (binary), and all-cause mortality (binary) from the internal representations learned by the digital twin of each trajectory until the first ICU day (Fig.~\ref{fig:R1}).
Across hospital sites, the embeddings achieve consistent, nontrivial predictive performance, indicating that they capture meaningful signal.
We also observe modest site-level structure in the embedding space, suggesting partial but not complete alignment across institutions.
In several tasks, including all-cause mortality prediction, the model achieved consistent performance across prediction horizons ranging from 28 to 180 days.
Overall, performance on these benchmark tasks is comparable to that of other clinical foundation models, suggesting that the learned representations are sufficiently informative to serve as the predictive backbone for downstream optimization via MPC.

\paragraph{Patient Forecasting.}
We evaluate the digital twin $p_\theta$ as a generative model of clinical time series using next-token prediction (NTP).
Performance is assessed via next-$m$ token set recall and token-level accuracy stratified by semantic type \citep{stein2023exposing}.
For multi-step forecasting, we measure whether true future tokens appear within the top-$i$ predictions over horizons $m \in \{5, 10\}$ (Table~\ref{tab:ntp_combined}, left).
We also report accuracy by token category (Table~\ref{tab:ntp_combined}, right), including medications, location/unit transitions, and laboratory, vital, and order events.
Medication tokens are predicted most accurately, reflecting their temporal persistence and repeated administration. In contrast, laboratory, vital, and order tokens exhibit lower accuracy, consistent with their higher temporal variability and the stochastic nature of short-term physiologic dynamics.

\begin{table}[h]
\centering
\caption{\small\textbf{Digital Twin Forecasting.} Next-token prediction (NTP) evaluation. {Left:} Overall performance (Set Recall, \%). {Right:} Accuracy stratified by token type.}
\label{tab:ntp_combined}
\vspace{-4mm}
\begin{minipage}{0.38\linewidth}
\centering
\small
% \textbf{(a) Overall}
\begin{tabular}{lcc}
\toprule
 & \multicolumn{2}{c}{Next-$m$ tokens} \\
\cmidrule(lr){2-3}
Top-$i$ & $m=5$ & $m=10$ \\
\midrule
10 & 86.2 & 82.8 \\
20 & 89.4 & 85.9 \\
50 & 92.9 & 89.9 \\
\bottomrule
\end{tabular}
\end{minipage}
% \hfill
\begin{minipage}{0.48\linewidth}
\centering
\small
% \textbf{(b) By token type}
\begin{tabular}{lccc}
\toprule
Token Type & Top-1 & Top-5 & Top-10 \\
\midrule
Medications & 54.8 & 74.0 & 82.7 \\
Location / unit & 31.4 & 58.7 & 66.0 \\
Labs, vitals \& orders & 27.2 & 54.3 & 65.0 \\
\bottomrule
\end{tabular}
\end{minipage}
\end{table}

\paragraph{Dose-Response Sensitivity.}
We evaluate whether the digital twin exhibits sensitivity to pharmacologic intervention tokens in line with Eq.~\ref{eq: action forced distribution}.
For each high-acuity patient (SOFA $\geq 6$), we construct counterfactual 24-hour trajectories by injecting repeated drug tokens into the observed context as if the drug were administered at a higher dose and then generate future tokens by sampling from the digital twin $p_\theta$.
The number of drug tokens is varied from 1 to 20, and estimated mortality risk is predicted using a task specific prediction head (see Table~\ref{tab:dose_response}).
Corticosteroid tokens induce a consistent, monotonic decrease in predicted mortality risk across increasing injection levels.
Antibiotics and vasopressors exhibit weaker and less monotonic responses, with more variability across dose levels.
These results suggest that the model does not respond uniformly to intervention tokens, but instead exhibits drug-specific variation.

The prediction and forecasting accuracy together with the intervention sensitivity satisfy empirical conditions for local controllability.
This supports the use of MPC to steer the learned digital twin.

\begin{table}[t]
\centering
\small
\setlength{\tabcolsep}{4pt}
\renewcommand{\arraystretch}{1.2}
\caption{\small\textbf{Dose-response sensitivity in the digital twin.} Predicted mortality under counterfactual rollouts with increasing numbers of injected intervention tokens. Values correspond to $P(\mathrm{mortality})$ from a frozen prediction head.}
\label{tab:dose_response}
\vspace{-3mm}
\begin{tabular}{lcccccccccc}
\toprule
\textbf{Dose tokens} & 1 & 2 & 4 & 6 & 8 & 10 & 12 & 15 & 18 & 20 \\
\midrule
Steroids    & 0.38 & 0.33 & 0.41 & 0.39 & 0.39 & 0.37 & 0.31 & 0.22 & 0.16 & 0.12 \\
Vasopressor & 0.49 & 0.56 & 0.58 & 0.66 & 0.68 & 0.73 & 0.67 & 0.62 & 0.54 & 0.52 \\
Antibiotics & 0.44 & 0.50 & 0.44 & 0.48 & 0.45 & 0.41 & 0.54 & 0.47 & 0.43 & 0.50 \\
\bottomrule
\end{tabular}
\end{table}

% \subsection{EHR-MPC Treatment Policy Optimization}\label{sec: exp 2}
% Using the learned digital twin $p_\theta$, we solve Eq.~\ref{eq: optimization} to optimize treatment recommendations. We evaluate policies under two complementary but imperfect frameworks. Off-policy evaluation reflects agreement with observed clinical behavior but suffers from high variance under policy shift. Simulation-based evaluation enables on-policy comparison under learned dynamics but depends on model fidelity. Hence, we interpret results jointly rather than relying on either in isolation.

\subsection{EHR-MPC Treatment Policy Optimization}\label{sec: exp 2}
Using the learned digital twin $p_\theta$, we solve Eq.~\ref{eq: optimization}  to optimize treatment policies.
Evaluating treatment policies learned from  observational EHR data presents significant statistical challenges.
Off-policy weighted importance sampling (WIS) estimates policy value from observed trajectories but exhibits high variance when the learned and observed policies differ substantially; WIS is included here to enable comparison with prior RL approaches for sepsis that rely on similar protocols.
Simulation-based evaluation enables on-policy rollout under the learned dynamics but introduces dependence on model fidelity.
Due to these fundamental limitations, we interpret results jointly across both evaluation frameworks, treating consistency between them as stronger evidence than either alone.

\begin{figure}[t]
    \centering
    \includegraphics[width=0.7\linewidth]{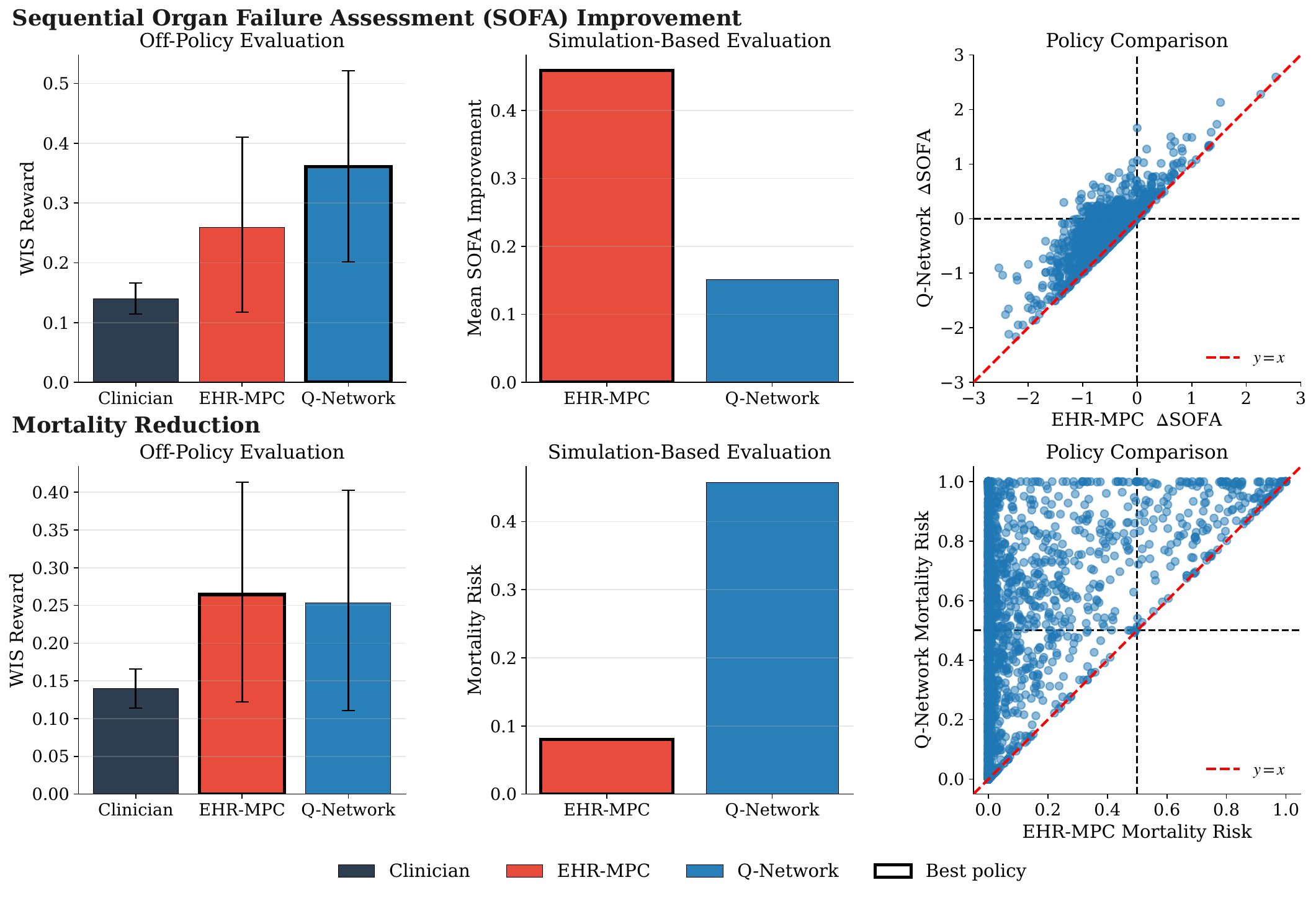}
    \vspace{-5mm}
    \caption{\small\textbf{Offline Evaluation of EHR-MPC Treatment Policies.} The learned EHR-MPC policy is evaluated with WIS (left) and simulated patient outcomes (middle, right). The improvement of patient outcomes according to EHR-MPC and Q-networks is shown per patient (right). The best policy designates the most advantageous treatment strategy identified among the tested approaches and according to each evaluation scheme.}
    \label{fig:R2}
\end{figure}

\paragraph{Off-Policy Evaluation via Importance Sampling.}
We evaluate \textsc{EHR-MPC} under two reward functions: (i) minimizing SOFA  score and (ii) minimizing mortality risk, comparing against a Q-network  baseline and the observed clinician policy using per-decision weighted importance sampling (WIS). Although $p_\theta$ is trained on the full ICU cohort, evaluation is restricted to sepsis patients. Across both objectives, \textsc{EHR-MPC} and the Q-network achieve higher estimated value than the clinician policy under WIS (Fig.~\ref{fig:R2}, left). Point estimates between \textsc{EHR-MPC} and the Q-network are similar with overlapping confidence intervals.

% \paragraph{Off Policy Evaluation via Importance Sampling.}
% We evaluate \textsc{EHR-MPC} under two reward functions: (i) minimizing SOFA score and (ii) minimizing mortality risk. We compare against a Q-network baseline and the observed clinician policy using per-decision weighted importance sampling (WIS), a standard but high-variance method for offline off-policy evaluation in sepsis \citep{komorowski2018artificial}. Although the digital twin $p_\theta$ is trained on the full ICU cohort, evaluation is restricted to sepsis patients.
% %
% Across both objectives, \textsc{EHR-MPC} and the Q-network achieve higher estimated value than the clinician policy under WIS (Fig.~\ref{fig:R2}, left). For SOFA minimization, the Q-network attains slightly higher point estimates than \textsc{EHR-MPC}, while confidence intervals substantially overlap. For mortality minimization, \textsc{EHR-MPC} yields a marginally higher point estimate, again with overlapping uncertainty. Overall, both learned policies outperform the clinician baseline, supporting the utility of $p_\theta$ as a predictive model and the quality of its learned representations.

% WIS exhibits high variance and sensitivity to distribution shift between learned and observed policies, which motivates the complementary simulation-based evaluation.

\paragraph{On-Policy Evaluation via Digital Twin Simulator.}
We assess learned policies using the digital twin as a simulator. Patient trajectories are initialized from observed tokens up to the first ICU day, after which each policy (\textsc{EHR-MPC} and Q-networks) selects interventions at daily intervals as the simulated patient evolves over time. Unlike WIS, which evaluates fixed logged trajectories under distribution shift, simulation-based evaluation enables on-policy rollouts under the learned digital twin. Across simulations, \textsc{EHR-MPC} consistently outperforms Q-network policies aimed to improve daily SOFA scores and decrease mortality risk, both in terms of average effect (Fig.~\ref{fig:R2}, center) and individual simulations (Fig.~\ref{fig:R2}, right). We examine the robustness of this result to model misspecification in the following experiment.

% \paragraph{On-Policy Evaluation via Digital Twin Simulator.}
% We assess learned policies using the digital twin as a simulator. Patient trajectories are initialized from observed tokens up to the first ICU day, after which each policy (\textsc{EHR-MPC} and Q-networks) selects interventions at daily intervals as the simulated patient evolves over time.
% %
% Unlike PDWIS, which evaluates fixed logged trajectories under distribution shift, simulation-based evaluation enables on-policy rollouts under the learned digital twin.

% Across simulations, \textsc{EHR-MPC} consistently outperforms Q-network policies aimed to improve daily SOFA scores and decrease mortality risk. This is shown both in terms of the average effect (Fig.~\ref{fig:R2}, center) and for individual simulations (Fig.~\ref{fig:R2}, right).
% %
% This is expected for two reasons. First, Q-networks are constrained by the behavior policy and learn only from observed trajectories, whereas \textsc{EHR-MPC} performs inference-time planning over simulated futures, allowing direct optimization of outcomes under the learned dynamics. Second, \textsc{EHR-MPC} explicitly exploits the digital twin $p_\theta$ during optimization in Eq.~\ref{eq: optimization}, which assumes access to a faithful model of patient dynamics. In practice, the second assumption is imperfect, as the learned simulator is only an approximation of true physiology. We examine the impact of this mismatch in the following experiment.

\begin{figure}
    \centering
    \includegraphics[width=\linewidth]{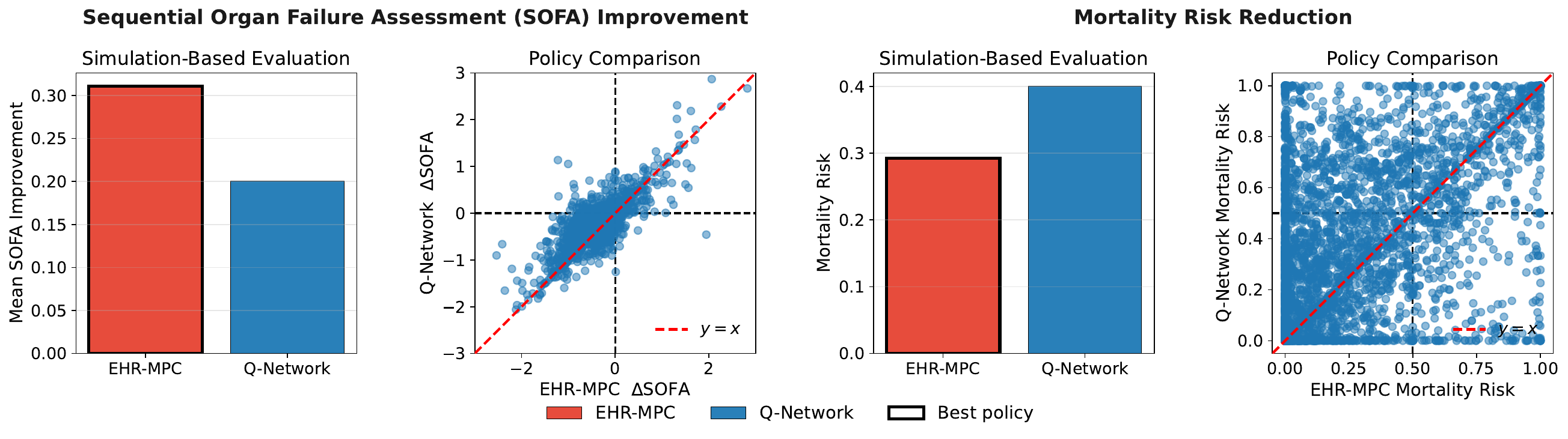}
    \vspace{-10mm}
    \caption{\small\textbf{MPC Robustness to Model Misalignment.} Performance of MPC and Q-networks are evaluated on the digital twin simulator when the MPC controller uses a less accurate version of the patient digital twin.}
    \label{fig:R3}
\end{figure}

\paragraph{Robustness to Model Misspecification.}
A potential concern with simulator-based evaluation is that \textsc{EHR-MPC} may trivially optimize the evaluation metric by exploiting the same dynamics model it plans with. To address this, we decouple the \emph{planning model} from the \emph{evaluation model} using checkpoints from different training epochs: MPC planning uses an early-epoch checkpoint (40\% of training epochs), while policy scoring and simulation use a later, more-converged checkpoint. The Q-network is also trained on the representations of the planning model used by MPC. This separates action selection from action evaluation, providing a more rigorous test of generalization under model mismatch.
Under this protocol, \textsc{EHR-MPC} continues to outperform the clinician policy with only modest performance degradation relative to the same-model baseline (Fig.~\ref{fig:R3}). Degradation is larger for mortality risk minimization than for SOFA, consistent with SOFA providing denser per-step rewards that are easier to optimize.

% \paragraph{Robustness to Model Misspecification.}
% A potential concern with simulator-based evaluation is that \textsc{EHR-MPC} may trivially optimize the evaluation metric by exploiting the same dynamics model it plans with.
% %
% To address this, we decouple the \emph{planning model} from the \emph{evaluation model} using checkpoints from different training epochs.
% %
% Concretely, MPC planning is performed using an early-epoch checkpoint (40\% of finetuning epochs), while all policy scoring and simulation is conducted using a later, more-converged checkpoint.
% %
% The Q-network remains trained on the state representations of the final converged model.
% %
% This separates the model used for action selection from the model used to evaluate those actions, providing a more rigorous test of how MPC generalizes.

% Under this cross-epoch evaluation protocol, \textsc{EHR-MPC} continues to outperform the clinician policy, and its performance degrades only modestly relative to the same-model baseline (Fig.~\ref{fig:R2}).
% %
% We observe that MPC degrades substantially more than mortality, when evaluating the effect on individual patients.
% %
% This is consistent with SOFA providing denser rewards that are easier for the dynamics model to track.
% %
% This demonstrates that the quality of MPC recommendations is not solely a product of within-model alignment, but reflects genuine improvement in predicted patient outcomes as assessed by a separate, more accurate patient digital twin.

\begin{figure}[t]
    \centering
    \includegraphics[width=0.9\linewidth]{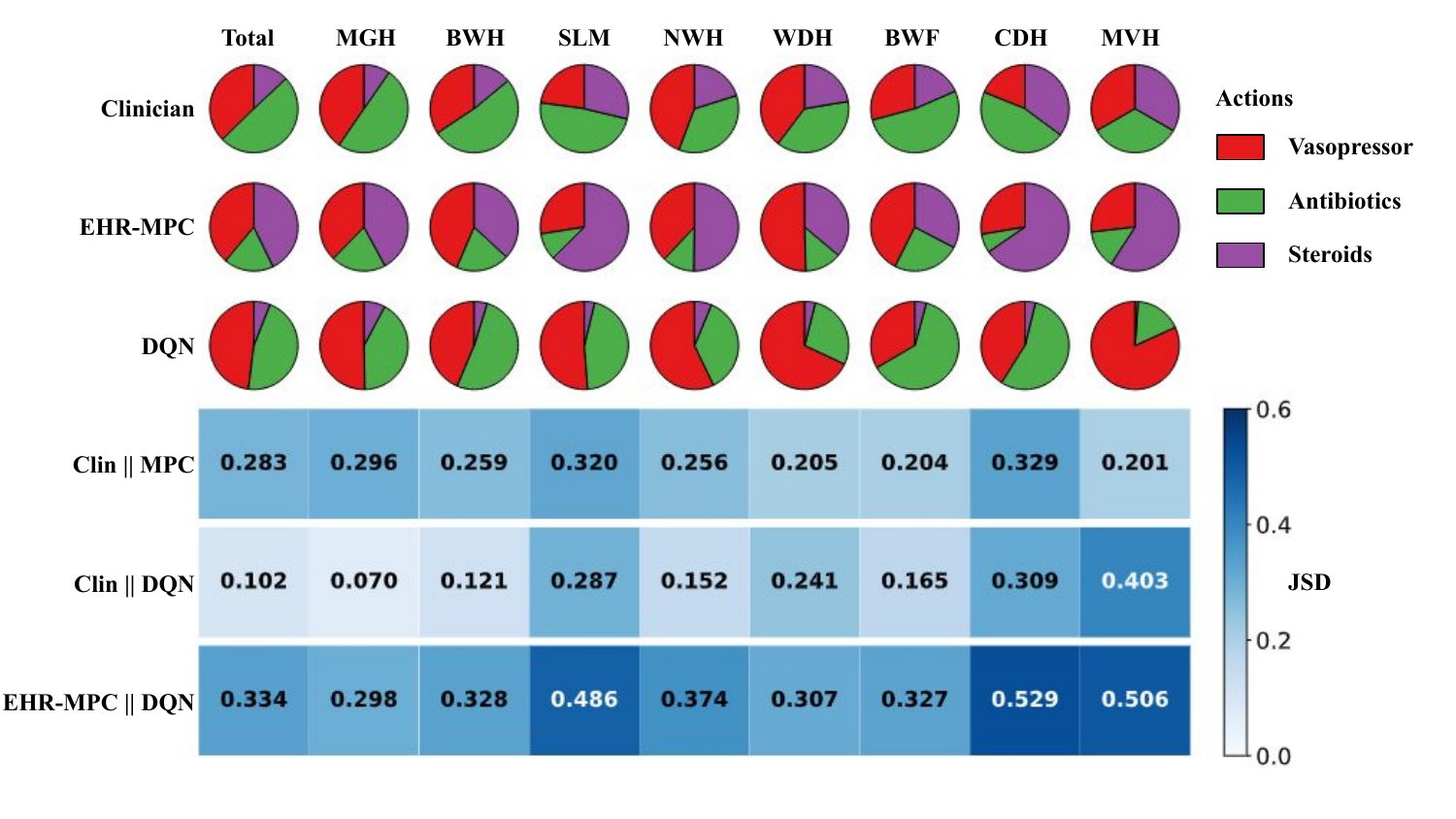}
    \vspace{-10mm}
    \caption{\small\textbf{Policy Divergence for Mortality Optimization.} (Top) The distribution of recommended drugs is shown for each of the three policies across locations. (Bottom) The JSD divergence between these distributions is shown.}
    \label{fig: policy variation}
\end{figure}

\paragraph{Treatment Policy Divergence.}
To complement the off-policy and simulation-based evaluations, we examine differences in action distributions between \textsc{EHR-MPC}, Q-networks, and the observed clinical policy (Fig.~\ref{fig: policy variation}, top).
The policies are stratified by location and recommendations of individual drugs.
We further quantify policy differences using Jensen-Shannon divergence (JSD) over action distributions across hospital sites (Fig.~\ref{fig: policy variation}, bottom). The Q-network policy remains close to the clinical policy across most sites, while \textsc{EHR-MPC} shows greater divergence from both. Finally, the clinical policy shows the greatest heterogeneity across sites, whereas both \textsc{EHR-MPC} and Q-network policies are more consistent across locations (Fig.~\ref{fig: apdx R1}). Analogous analyses for SOFA score reduction are provided in Figs.~\ref{fig: apdx R2} and~\ref{fig: apdx R3}.

% Why EHR-MPC outperforms Q-networks in simulation and what that means structurally
% Why the WIS and simulation results together constitute meaningful evidence despite individual weaknesses
% The clinical trial alignment as external validation
% Model fidelity as a limitation

\section{Discussion}
EHR-MPC frames sepsis treatment optimization as an inference-time planning problem over a learned generative model of patient dynamics.
The framework separates two distinct optimization problems of (i) modeling how patients evolve under interventions and (ii) selecting interventions that optimize clinical goals.
Previous RL for sepsis addressed these tasks jointly to learn a fixed policy.
Once $p_\theta$ is learned, EHR-MPC determines treatment strategies that are flexible to new clinical objectives without retraining, providing a robust and generalizable tool for decision-making at inference time.

\paragraph{Interpreting Policy Evaluations.}
Based on the WIS, the EHR-MPC's comparable performance relative to Q-networks demonstrates the decoupled framework for treatment optimization can achieve comparable performance to RL methods that study sepsis treatment optimization.
Moreover, the increased divergence between EHR-MPC and the clinician while maintaining comparable performance with the Q-networks, which have policies more similar to the clinicians, indicate that EHR-MPC overcomes the penalty of WIS for diverging from observed data.
This is an important property for searching and optimizing the treatment space.
Based on the simulations, \textsc{EHR-MPC} consistently outperforms Q-networks on both SOFA improvement and mortality reduction, and this advantage persists under the model misspecification protocol, where the MPC controller planned over a poorly identified simulation model.
This robustness rules out the concern that \textsc{EHR-MPC} is merely exploiting a shared dynamics model, and supports the interpretation that online planning provides a robust framework for treatment optimization.
Across both evaluation schemes, which are standard in RL and MPC respectively, the results support the use of EHR-MPC as a framework for inference-time treatment optimization.

Examining the action distributions provides further context for these results.
Regarding specific treatment recommendations, the Q-network reduces corticosteroid utilization relative to clinical practice, consistent with prior RL studies \citep{bologheanu2023development}.
In contrast, \textsc{EHR-MPC} recommends increased steroid use when optimized for mortality reduction, more consistent with clinical trial evidence that corticosteroids reduce 28-day mortality in sepsis patients \citep{annane2025corticosteroids}.
The site-level variation shows that both learned policies are more consistent than the clinical policy.

Taken together, \textsc{EHR-MPC} is supported both empirically---by strong simulation performance and off-policy estimates---and clinically, by producing treatment recommendations that better align with the broader evidence base than competing algorithms.

\paragraph{Capabilities enabled by inference-time control.}
By decoupling patient dynamics from treatment optimization, EHR-MPC supports several capabilities that are difficult to achieve with RL models.
First, treatment strategies are computed via inference-time planning, allowing decisions to be adapted to patient state and evolving clinical objectives without retraining.
Unlike many medical and engineering applications where low-latency control is essential, clinical decision-making occurs on substantially longer timescales than the inference-time computation required by EHR-MPC \citep{ching2013real}.
While the MPC planning stage may require seconds to minutes to generate a recommendation, clinicians make treatment decisions for interventions such as antibiotics, corticosteroids, and other medications over hours-to-days timescales \citep{prescott2026surviving}.
Second, a single learned dynamics model can support multiple clinical objectives, enabling trade-offs between competing goals.
This allows the EHR-MPC framework to support many other EHR trajectory models beyond the transformer-based architecture studied here \citep{zeng2025trajsurv, zeng2025traj, shmatko2025learning}.
Third, safety and feasibility constraints can be incorporated directly during planning, allowing candidate interventions to be filtered or penalized without requiring changes to the underlying model.
Finally, the framework explicitly models and predicts patient trajectories under interventions, providing clinicians with interpretable forecasts of patient state rather than only scalar reward estimates.
Together, these capabilities position EHR-MPC as a flexible decision-support framework that aligns with the requirements of clinical practice.

\paragraph{Limitations.}
The primary limitation of this work is the fidelity and validation of the learned digital twin.
While $p_\theta$ is trained on intervention-aware patient trajectories, the training process does not provide guarantees of correct system identification.
This reflects a well-known challenge in system identification and control \citep{ljung1998system, bemporad2007robust}, and more broadly in off-policy evaluation for sequential decision-making in healthcare \citep{shalit2017estimating, oberst2019counterfactual}.
The model is trained on observational data confounded by treatment indication, which may result in hallucinated or unreliable patient trajectories when forecasting outcomes under treatment regimes that are not sufficiently represented in the training data \citep{seedat2022continuous}.
Uncertainty quantification, observability analysis, and improving the reliability of learned patient dynamics, as well as validating their behavior under interventions, remain important and open problems in data-driven clinical modeling and control.

Second, offline policy evaluation remains a substantial challenge for validating this work along with other RL for sepsis optimization.
Off-policy estimators such as WIS exhibit high variance and sensitivity to support mismatch between policies, specifically when evaluating policies that diverge from clinician behavior \citep{precup2000eligibility, gottesman2018evaluating}.
In contrast, simulation-based evaluation alleviates these statistical issues by enabling rollout under the learned dynamics, but introduces dependence on model accuracy.
Neither evaluation framework alone is sufficient, but their consistency provides stronger evidence than either in isolation.
While EHR-MPC enables simulation-based, on-policy evaluation, substantial work remains to improve the fidelity, generalizability, and validation of such simulations.

From a computational standpoint, this work adopts standard architectures and evaluation schemes to isolate the effect of the proposed framework.
However, each component, including the generative model, the MPC planner, and off-policy evaluation, could be further optimized.
While the results demonstrate the effectiveness of EHR-MPC, performance may be improved through more advanced architectures and more extensive hyperparameter tuning.
Nonetheless, these results establish the utility of EHR-MPC under standard modeling choices, suggesting opportunities for improvement through advances in model architecture and optimization that are orthogonal to the proposed framework.

Ultimately, the value of decoupling dynamics from policy optimization depends on how faithfully the digital twin captures patient physiology under intervention.
Improving digital twin fidelity, prospectively validating learned dynamics, and developing principled evaluation frameworks for simulation-based policies remain the central open problems in translating this class of methods into clinical practice.

\vspace{5mm}
\noindent
\textbf{Acknowledgments.}
The authors wish to thank the Eric and Wendy  Schmidt Center at the Broad Institute of MIT and Harvard and the Center for Integrated Solutions for Infectious Diseases (CISID) for supporting this work.

\bibliography{references}

\newpage
\appendix

% \textcolor{blue}{\bf I am filling in additional architecture, compute, hyperparameter details tonight etc.}

\section{Cohort and Data Description}\label{sec: cohort and data}

\subsection{Patient Categorization and Clinical Rewards}

\paragraph{Disease definitions.}
Sepsis and related comorbidities were identified using ICD-10 codes as listed in Table~\ref{tab:ICD}.

\begin{table}[h]
    \centering
    \begin{tabular}{l|l}
    \hline
    \textbf{Diagnosis}&\textbf{ICD-10 Code}\\
    \hline
    Sepsis              & A40, A41, R65.2\\
    Septic shock        & R65.21\\
    Respiratory failure & J96, J80\\
    AKI                 & N17\\
    CKD                 & N18\\
    Hepatic failure     & K72\\
    Heart failure       & I50\\
    Coagulopathy / DIC  & D65, D68\\
    Encephalopathy      & G93.4, F05\\
    Pneumonia           & J18, J15\\
    Shock (other)       & R57.0, R57.1, R57.8, R57.9\\
    \hline
    \end{tabular}
    \caption{ICD-10 codes used to define clinical conditions in the study cohort.}
    \label{tab:ICD}
\end{table}

\paragraph{Mortality.}
In-hospital mortality was defined as a documented death during the index hospital encounter or within 72 hours of discharge.
Mortality labels were extracted from the Epic EHR system's outcome and encounter tables.

\paragraph{SOFA score.}
The Sequential Organ Failure Assessment (SOFA) score was computed from raw EHR flowsheet and laboratory data across six organ systems (Table~\ref{tab:sofa_rules}).
Each patient's stay was partitioned into non-overlapping 24-hour windows anchored at first event; patients with fewer than 24 hours of data were excluded.
Within each window, the worst (most abnormal) value was used for each component, except PaO$_2$ and platelets where the lowest observed value is the worst.
Subscores follow standard 0--4 SOFA thresholds; the total is their sum over non-missing components.

Where arterial PaO$_2$ was unavailable, it was estimated from SpO$_2$ via a linear approximation of the oxygen--haemoglobin dissociation curve: $\widehat{\text{PaO}}_2 = 3(\text{SpO}_2 - 90) + 60$\,mmHg, valid for SpO$_2\in[75,100]\%$.
FiO$_2$ was resolved by hierarchy: (i)~directly measured, (ii)~estimated from O$_2$ flow rate as $20 + 4{\times}\dot{V}\,(\text{L/min})$ capped at 60\%, or (iii)~assumed 21\%.
Mechanical ventilation was inferred from O$_2$-device flowsheet values containing ``vent''; SOFA respiratory subscores of 3--4 require ventilation (else capped at 2).
Vasopressor detection used a fixed-score of 2 regardless of dose, as reliable infusion-rate data were unavailable.

$\Delta\text{SOFA}$ was computed component-wise between consecutive windows and summed only over components non-missing in \emph{both} windows, avoiding artifacts from missing data:
\begin{equation}
  \Delta\text{SOFA}^{(t)} = \textstyle\sum_{k\in\mathcal{K}_t}\!\bigl(s_k^{(t+1)} - s_k^{(t)}\bigr),
  \label{eq:delta_sofa}
\end{equation}
where $\mathcal{K}_t$ is the set of components with non-missing scores in both window $t$ and $t{+}1$.

\begin{table}[t]
\centering
\caption{\textbf{SOFA computation rules per 24-hour window.} All EHR field patterns are case-insensitive regular expressions matched on flowsheet event names. ``Agg.''\ denotes within-window aggregation. $^\dagger$Scores 3--4 require concurrent mechanical ventilation; otherwise subscore capped at 2.}
\label{tab:sofa_rules}
\footnotesize % Slightly smaller than \small for better fit
\setlength{\tabcolsep}{3pt}
\renewcommand{\arraystretch}{1.4}
% \begin{tabular}{\textwidth}{@{}p{2.2cm} X c p{2.8cm} X @{}}
\begin{tabular}{@{}
p{2.2cm}
p{4.3cm}
c
p{2.8cm}
p{5.0cm}
@{}}
% \begin{tabularx}{\textwidth}{@{}l X c p{2.8cm} X @{}}
\toprule
\textbf{Component} & \textbf{EHR Field Pattern(s)} & \textbf{Agg.} & \textbf{Thresholds (Score: Range)} & \textbf{Implementation Notes} \\ \midrule
\textbf{Respiratory} \newline {\scriptsize PaO$_2$/FiO$_2$} & 
    \texttt{\^{}PO2} (excl. venous) \newline 
    \texttt{\^{}SpO2\textbackslash{}s*|/|} \newline 
    \texttt{\^{}FIO2\textbackslash{}s*|/|} \newline 
    \texttt{O2 Flow Rate} & 
    Min/Max & 
    \textbf{0:} $\geq$400 \newline 
    \textbf{1:} [300, 400) \newline 
    \textbf{2:} [200, 300) \newline 
    \textbf{3:} [100, 200)$^\dagger$ \newline 
    \textbf{4:} $<$100$^\dagger$ & 
    If PaO$_2$ absent, impute from SpO$_2$: $\widehat{\mathrm{PaO}}_2 = 3(\mathrm{SpO}_2 - 90) + 60$ mmHg. FiO$_2$ hierarchy: measured $\to$ flow estimate ($20+4 \times$ L/min) $\to$ 21\%.  \\ \midrule

\textbf{Coagulation} \newline {\scriptsize Platelets $10^3/\mu$L} & 
    \texttt{\^{}PLT\$} & 
    Min & 
    \textbf{0:} $\geq$150 \newline 
    \textbf{1:} [100, 150) \newline 
    \textbf{2:} [50, 100) \newline 
    \textbf{3:} [20, 50) \newline 
    \textbf{4:} $<$20 & 
    \\ \midrule

\textbf{Hepatic} \newline {\scriptsize Bilirubin mg/dL} & 
    \texttt{\^{}Total Bilirubin\$} \newline 
    \texttt{\^{}Bilirubin,\textbackslash{}s*Total} & 
    Max & 
    \textbf{0:} $\leq$1.2 \newline 
    \textbf{1:} (1.2, 2] \newline 
    \textbf{2:} (2, 6] \newline 
    \textbf{3:} (6, 12] \newline 
    \textbf{4:} $>$12 & 
    \\ \midrule

\textbf{Cardio} \newline {\scriptsize MAP mmHg} & 
    \texttt{art.*\textbackslash{}bmap\textbackslash{}b} \newline 
    Fallback: \texttt{\^{}MAP\textbackslash{}b} \newline 
    Medication names & 
    Min & 
    \textbf{0:} MAP $\geq$70 \newline 
    \textbf{1:} MAP $<$70 \newline 
    \textbf{2-4:} Dependent on vasoactive medications dose
    & 
    ART preferred over NIBP  \\ \midrule

\textbf{Neuro} \newline {\scriptsize GCS} & 
    \texttt{Glasgow Coma Scale} & 
    Min & 
    \textbf{0:} 15 \newline 
    \textbf{1:} [13, 15) \newline 
    \textbf{2:} [10, 13) \newline 
    \textbf{3:} [6, 10) \newline 
    \textbf{4:} $<$6 & 
    Missing window yields NaN rather than imputing \\ \midrule

\textbf{Renal} \newline {Creatinine mg/dL} & 
    \texttt{\^{}Creatinine\$} \newline 
    \texttt{\^{}Creatinine\textbackslash{}s*Whole Bld\$} & 
    Max & 
    \textbf{0:} $\leq$1.2 \newline 
    \textbf{1:} (1.2, 2.0] \newline 
    \textbf{2:} (2.0, 3.5] \newline 
    \textbf{3:} (3.5, 5.0] \newline 
    \textbf{4:} $>$5.0 & \\
\bottomrule
\end{tabular}
\end{table}

\newpage

\subsection{Patient Tokenization}\label{apdx: tokenization}
Tables \ref{tab:tokens_global} and \ref{tab:tokens_by_hospital} summarize the tokenization scheme.

% Table A — Global tokenization statistics
\begin{table}[htbp]
\centering
\caption{\textbf{Tokenization statistics.} Content tokens exclude special tokens: \texttt{[BOS], [EOS], [PAD], [UNK], [MASK], [4 hours]}.}
\label{tab:tokens_global}
\small
\begin{tabular}{l r l}
\toprule
\textbf{Statistic} & \textbf{Value} & \textbf{Unit} \\
\midrule
\multicolumn{3}{l}{\textit{Vocabulary}} \\
\quad Patients & 36,930 & people \\
\quad Vocabulary size & 96,268 & tokens \\
\quad Content tokens & 469,272,491 & tokens \\

\multicolumn{3}{l}{\textit{Token type breakdown (content tokens; cohort total)}} \\
\quad Observation (vital / lab) & 423,557,340 (90.3\%) & tokens \\
\quad Medication & 24,030,597 (5.1\%) & tokens \\
\quad Procedure / order & 17,277,634 (3.7\%) & tokens \\
\quad Administrative & 4,306,243 (0.9\%) & tokens \\

\multicolumn{3}{l}{\textit{Sequence statistics per patient (median, IQR)}} \\
\quad Sequence length & 20232 (12007--28978) & tokens \\
\quad Unique tokens & 1058 (756--1533) & tokens \\
% \quad Unique event-name prefixes & 454 (359--565) & names \\

\multicolumn{3}{l}{\textit{Per-patient content-token fractions (median, IQR)}} \\
\quad Observation fraction & 88.6 (84.6--92.8) & \% \\
\quad Medication fraction & 4.6 (2.9--6.5) & \% \\
\quad Procedure fraction & 4.4 (2.9--6.5) & \% \\
\quad Administrative fraction & 1.0 (0.4--2.5) & \% \\

\multicolumn{3}{l}{\textit{Complexity and quality}} \\
\quad Token entropy & 3.86 (2.11--5.50) & nats \\
\quad Vocab coverage (unique / length) & 5.6 (4.0--8.7) & \% \\
\quad UNK rate & 0.01 (0.00--0.03) & \% \\
\bottomrule
\end{tabular}
\end{table}

% Table B — Per-hospital tokenization statistics
% \begin{landscape}
\begin{table}[htbp]
\centering
\caption{\textbf{Per-hospital tokenization statistics.}}
\label{tab:tokens_by_hospital}
\fontsize{7.8}{10}\selectfont
\begin{tabular}{l>{\columncolor[gray]{0.92}}rrrrrrrrr}
\toprule
%\textbf{Statistic} & \textbf{Overall} & \textbf{MGH} & \textbf{BWH} & \textbf{SLM} & \textbf{NWH} & \textbf{WDH} & \textbf{BWF} & \textbf{CDH} & \textbf{MVH} \\
\textbf{Statistic} & \textbf{Total}&\textbf{MGH}&\textbf{BWH}&\textbf{SLM}&\textbf{NWH}&\textbf{WDH}&\textbf{BWF}&\textbf{CDH}&\textbf{MVH}\\
% \textbf{Overall} & \textbf{A} & \textbf{B} & \textbf{C} & \textbf{D} & \textbf{E} & \textbf{F} & \textbf{G} & \textbf{H} \\
\midrule
\multicolumn{10}{l}{\textit{Sequence statistics (median)}} \\[2pt]
Patients, $n$ & 36,930 & 13,651 & 12,291 & 3,541 & 2,256 & 1,849 & 1,465 & 1,358 & 519 \\
\quad Sequence length (tokens) & 20232 & 22878 & 20940 & 19568 & 19827 & 11155 & 20165 & 14535 & 16485 \\
\quad Unique tokens per patient & 1058 & 1258 & 1191 & 840 & 853 & 771 & 820 & 772 & 490 \\

\multicolumn{10}{l}{\textit{Content-token fractions (\%)}} \\[2pt]
\quad Observation (vital / lab) & 88.6 & 90.2 & 89.9 & 86.2 & 85.6 & 86.5 & 85.7 & 84.8 & 83.1 \\
\quad Medication & 4.6 & 4.4 & 4.5 & 4.5 & 5.3 & 4.4 & 5.1 & 5.4 & 3.8 \\
\quad Procedure / order & 4.4 & 3.6 & 4.0 & 5.4 & 6.0 & 6.2 & 6.1 & 6.5 & 6.8 \\
\quad Administrative & 1.0 & 0.6 & 0.7 & 3.1 & 2.1 & 2.0 & 2.0 & 2.4 & 4.8 \\

\multicolumn{10}{l}{\textit{Complexity and quality (median)}} \\[2pt]
\quad Token entropy (nats) & 3.86 & 4.48 & 4.59 & 2.12 & 2.16 & 3.10 & 2.15 & 2.32 & 1.03 \\
\quad UNK rate (\%) & 0.01 & 0.01 & 0.01 & 0.02 & 0.02 & 0.02 & 0.02 & 0.03 & 0.03 \\
\bottomrule
\end{tabular}
\end{table}

\newpage
\section{Experimental Details}\label{apdx: experiment details}
\subsection{Digital Twin Architecture}

The patient digital twin is a decoder-only transformer with the following architecture:

\begin{itemize}
  \item \textbf{Model family:} GPT-2-style causal language model trained for next token prediction on the tokenized EHR data and finetuned task heads for predicting mortality risk and daily changes in SOFA.
  \item \textbf{Parameters:} 74.8M total (backbone).
  \item \textbf{Embedding dimension:} $d = 512$.
  \item \textbf{Vocabulary:} consists of discretized vitals, labs, medications, procedures, administrative events, demographics, and special tokens (\texttt{[BOS]}, \texttt{[EOS]}, \texttt{[PAD]}, \texttt{[UNK]}, \texttt{[4\,hour]}).
  \item \textbf{Context length:} 512 tokens (block size).
\end{itemize}
The base transformer model was trained on 7 NVIDIA A6000 GPUs for approximately 18 hours.

Patient trajectories are long, exceeding the model’s context window. To obtain patient trajectory-level representations for outcome prediction, we aggregate representations across sequential, non-overlapping windows spanning the trajectory. This design aligns with the approximately Markovian nature of short-horizon ICU dynamics, consistent with previous sepsis simulators \citep{bock2022superhuman}.

\paragraph{Evaluating Next Token Prediction.}
Standard top-$i$ accuracy measures whether the single true next token appears in the model's top-$i$ predictions.
Set Recall generalizes this to multi-step forecasting horizons.

Given a context prefix $x_{1:k}$, define the \emph{ground-truth set} $\mathcal{T}_m = \{x_{k+1}, x_{k+2}, \ldots, x_{k+m}\}$ as the set of distinct token identities observed in the next $m$ positions.
Let $\hat{\mathcal{T}}_i$ be the set of token identities in the model's top-$i$ predictions from the last-position logits:
\begin{equation}
    \hat{\mathcal{T}}_i = \left\{\, \text{id}_j \;\middle|\; j \in \mathrm{argtop}_i\, p_\theta(x_{k+1} \mid x_{1:k})\,\right\}.
\end{equation}
Set Recall at budget $i$ over horizon $m$ is:
\begin{equation}
    \mathrm{SR}@(i, m) = \frac{|\mathcal{T}_m \cap \hat{\mathcal{T}}_i|}{|\mathcal{T}_m|}.
\end{equation}
For each patient in the held-out set, we sample random positions as context endpoints and collect both the immediate next token (for top-$i$ accuracy) and the next-$m$ token set (for Set Recall).

\subsection{Outcome Prediction Heads}

\paragraph{SOFA delta head.}
A two-layer MLP with architecture $512 \to 256 \to 7$, predicting 24-hour per-component SOFA deltas. Trained with mean squared error loss on frozen backbone representations. The seven outputs correspond to total SOFA and six organ-system components (respiratory, coagulation, liver, cardiovascular, CNS, renal).

\paragraph{Mortality prediction head.}
A three-layer MLP with architecture
$\text{LayerNorm}(512) \to \text{Linear}(512, 256) \to \text{GELU} \to \text{Dropout}(0.2) \to \text{Linear}(256, 128) \to \text{GELU} \to \text{Dropout}(0.2) \to \text{Linear}(128, 1)$,
predicting log-odds of in-hospital mortality.
Trained with binary cross-entropy loss with class-weight balancing (positive rate: 38.5\%) for 100 epochs with early stopping (patience 20), learning rate $10^{-3}$, batch size 256, using Adam optimizer.
Achieves AUROC = 0.9352 and AUPRC = 0.9142 on held-out patients.

Separate heads are trained for different backbone checkpoints (early and late fine-tuning epochs) to support cross-epoch evaluation.

\subsection{MPC Optimization Procedure}\label{si:mpc-details}
The MPC planning algorithm is outlined in Algorithm~\ref{alg:ehr-mpc}.
At each decision point, four intervention options are considered: giving a combination of vasopressor, antibiotics, or steroids, or the option to not give one of these medications.
Up to 5 representative drug tokens are considered per drug class.
The final hidden state (last-token representation) of the transformer after rollout is extracted and passed to the outcome prediction head.
The action minimizing the predicted outcome score is selected.

The autoregressive generation uses a greedy (argmax) decoding strategy, generating future tokens token-by-token until six \texttt{[4\,hour]} tokens have been produced (approximately 24 hours of simulated patient state).

\begin{algorithm}[h]
\caption{EHR-MPC Planning via Token-Forced Rollouts}
\label{alg:ehr-mpc}
\begin{algorithmic}[1]
\REQUIRE Current patient state $x_{1:k_t}$, digital twin $p_\theta$, reward function $R$, planning horizon $h$, number of candidates $N$
\FOR{$i = 1$ to $N$}
    \STATE Sample candidate action sequence $a_{k_t:k_{t+h}}^{(i)}$
    \STATE Initialize trajectory $\hat{x}_{1:k_t}^{(i)} \leftarrow x_{1:k_t}$
    \FOR{$\tau = k_t$ to $k_{t+h}$}
        \STATE Force action token $a_{\tau}^{(i)}$ into model input
        \STATE Sample next token:
        \[
        \hat{x}_{k_{\tau+1}}^{(i)} \sim p_\theta\big(x_{k_{\tau+1}} \mid \hat{x}_{1:k_{\tau}}^{(i)}, a_{k_t:k_\tau}^{(i)}\big)
        \]
    \ENDFOR
    \STATE Compute trajectory reward:
    \[
    J^{(i)} \leftarrow R\big(\hat{x}_{k_t:k_{t+h}}^{(i)}\big)
    \]
\ENDFOR
\STATE Select best sequence:
\[
i^* \leftarrow \arg\max_i J^{(i)}
\]
\STATE \textbf{return} first action $a_t^{(i^*)}$
\end{algorithmic}
\end{algorithm}

\subsection{Per-Decision Weighted Importance Sampling (PDWIS)}

The behavior policy $\mu(a \mid s)$ is estimated by fitting a multinomial logistic regression classifier on the concatenated last-token representations of all training-set patient decision steps, predicting which of the four action classes was taken by the clinician.
At each evaluation step $t$ for trajectory $i$, the per-step IS ratio is
\begin{equation}
  \rho_t^{(i)} = \frac{\pi(a_t^{(i)} \mid s_t^{(i)})}{\mu(a_t^{(i)} \mid s_t^{(i)})},
\end{equation}
where $\pi$ is the evaluation policy (softmax over predicted scores, temperature $\tau = 0.5$).
Cumulative weights are clipped at $c_\text{max} = 10$ to reduce variance:
\begin{equation}
  w_t^{(i)} = \prod_{\tau \leq t} \min\!\left(\rho_\tau^{(i)},\, c_\text{max}\right).
\end{equation}
The PDWIS estimate is
\begin{equation}
  \hat{V}^\pi = \frac{\sum_i \sum_t w_t^{(i)}\, r_t^{(i)}}{\sum_i \sum_t w_t^{(i)}},
\end{equation}
where $r_t^{(i)} = -\Delta\text{SOFA}_t^{(i)}$ (positive when SOFA improves) is taken from the observed EHR data.
% Bootstrap confidence intervals ($B = 500$, 95\%) are computed by resampling patient trajectories.

\subsection{Reinforcement Learning Baselines}
\label{sec:rl_baselines}
To situate our framework against existing methods, we compare against
Q-network-based policies trained on the same frozen patient representations
$\mathbf{h}_p$.
A Q-network $Q_\psi : \mathbb{R}^{512} \times \mathcal{A} \to \mathbb{R}$
estimates the expected discounted return of each (state, action) pair under
a fixed reward function $r$, and is trained to minimize the Bellman
residual:
\begin{equation}
  \mathcal{L}_{\mathrm{Q}}(\psi) =
  \mathbb{E}\!\left[
    \left(
      r(p, a) + \gamma \max_{a'} Q_{\bar\psi}(\mathbf{h}_{p'}, a')
      - Q_\psi(\mathbf{h}_p, a)
    \right)^{\!2}
  \right],
  \label{eq:q_loss}
\end{equation}
where $\gamma$ is the discount factor and $Q_{\bar\psi}$ is a periodically
updated target network.
Q-networks constitute the canonical fused policy-dynamics baseline: the
value function implicitly encodes both patient dynamics and the clinical
objective, and adapting to a new reward function requires full retraining
of $Q_\psi$.
We train separate Q-networks for each reward function considered in our
experiments.
Each Q-network was trained on a single A6000 GPU for up to 300 epochs with a learning rate of $10^{-3}.$

% and then additional models including the Q-networks and outcome prediction heads were trained for up to 300 epochs on single GPUs.

\subsection{Additional Results}
\begin{figure}[h]
    \centering
    \includegraphics[width=\linewidth]{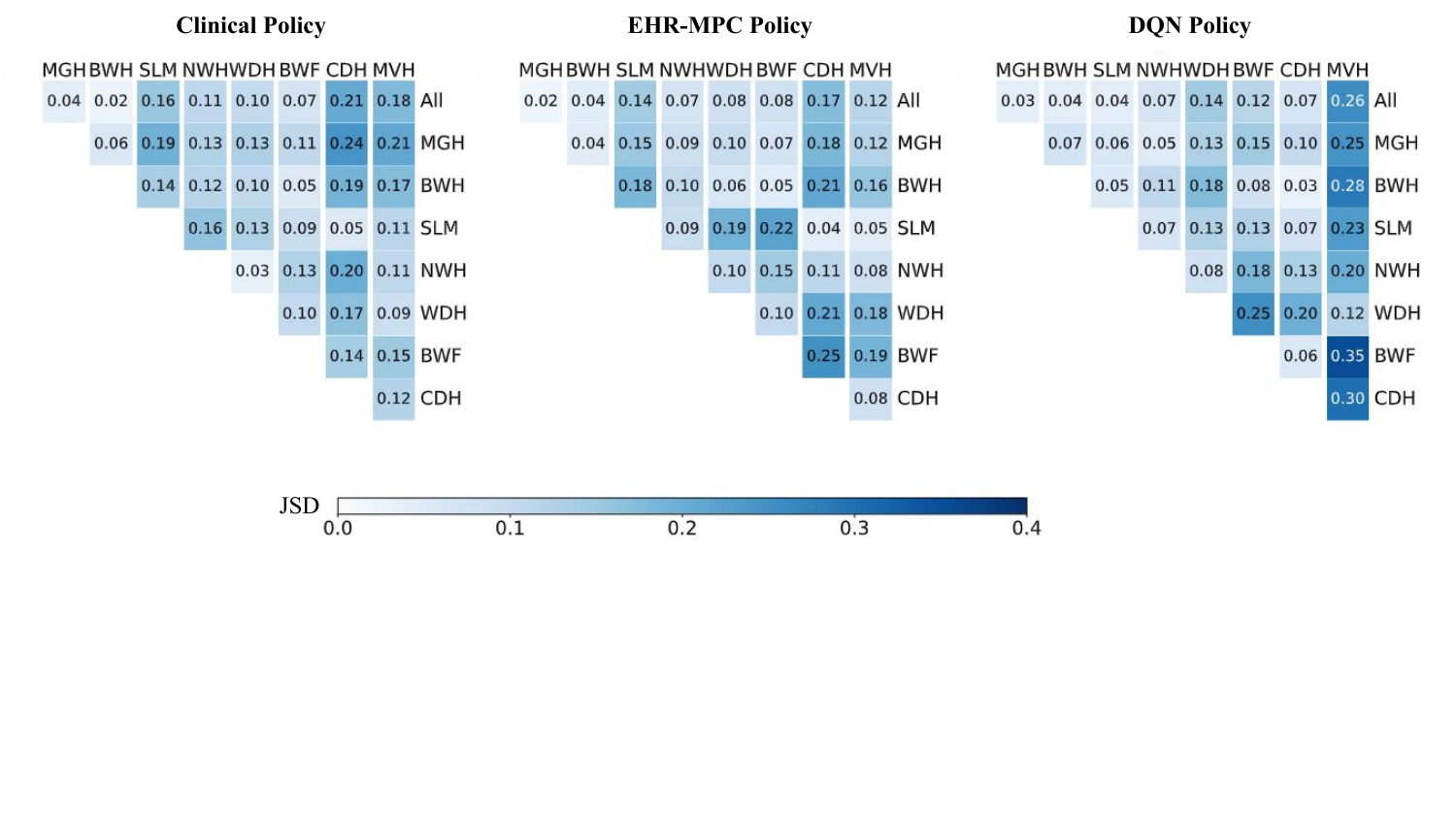}
    \vspace{-35mm}
    \caption{\textbf{Policy Variation for Mortality Risk Minimization.}}
    \label{fig: apdx R1}
\end{figure}

\begin{figure}[h]
    \centering
    \includegraphics[width=\linewidth]{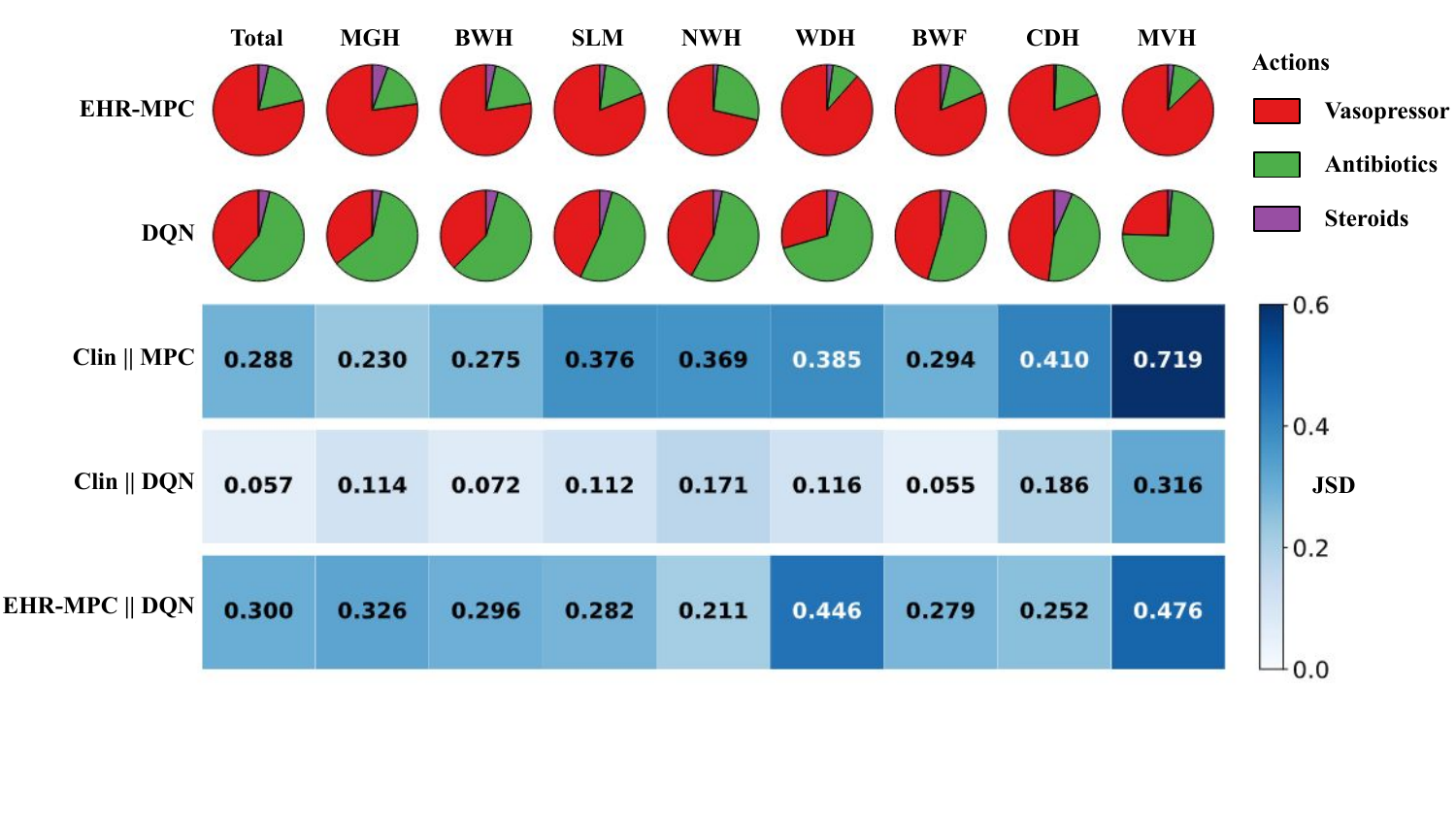}
    \vspace{-20mm}
    \caption{\textbf{Policy Distributions and Variation for SOFA Minimization.}}
    \label{fig: apdx R2}
\end{figure}

\begin{figure}[h]
    \centering
    \includegraphics[width=\linewidth]{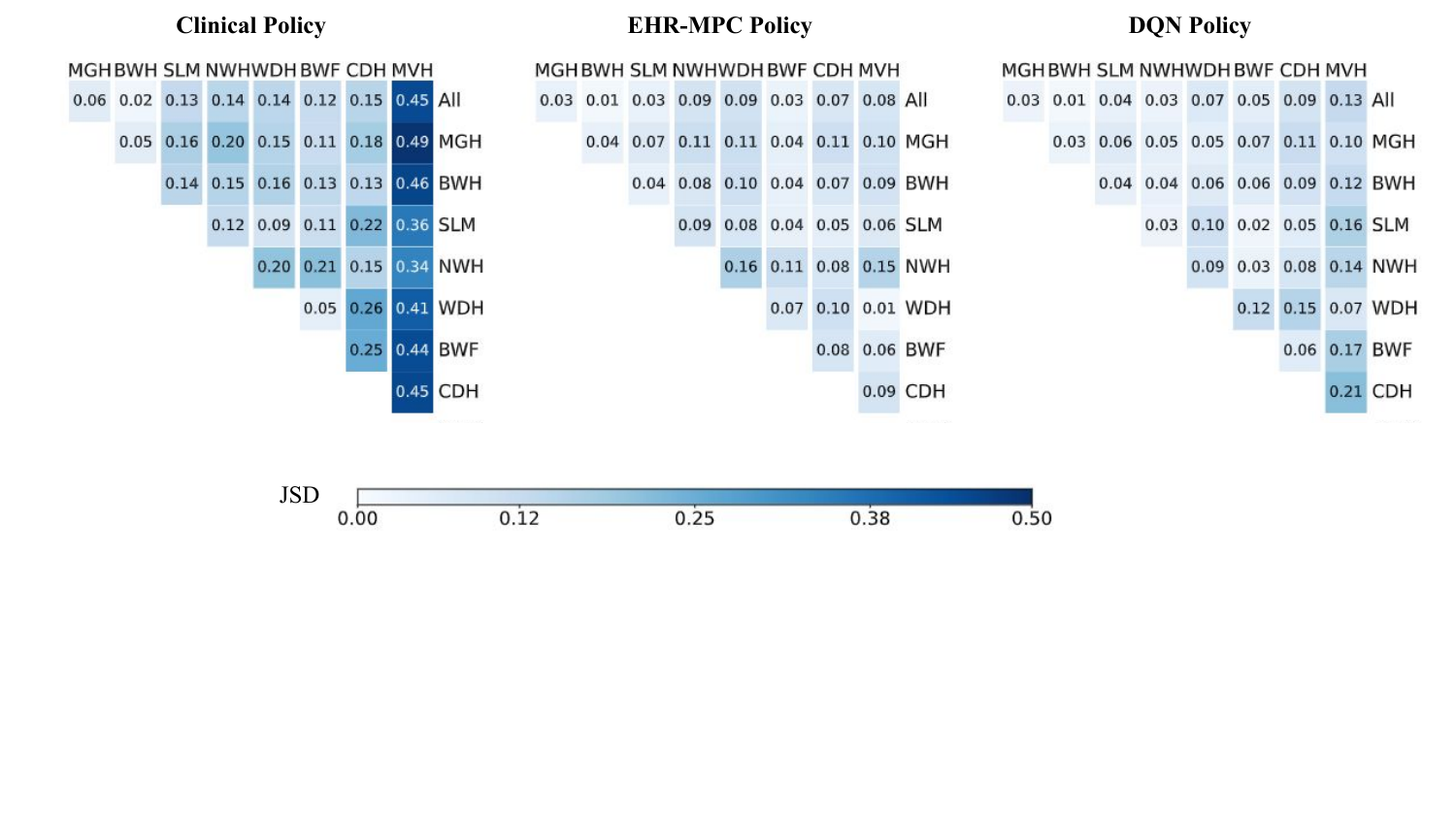}
    \vspace{-37mm}
    \caption{\textbf{Policy Variation for SOFA Minimization.}}
    \label{fig: apdx R3}
\end{figure}

Figures \ref{fig: apdx R1}-\ref{fig: apdx R3} illustrate policy distributions across mortality and SOFA rewards and locations.

%% appendix_v5.tex — Full Technical Appendix for EHR-MPC (MLHC 2026)
%% Drop this file into the same directory as main_v5.tex and add
%%   \input{appendix_v5}
%% immediately before \end{document}.  It picks up all packages from the main file.

% \newpage\clearpage
% \section{Extended Experimental Details}\label{apdx:extended}

\end{document}